\newcommand{\vggfigcut}{}
\newcommand{\vggfigcnp}{}

\RequirePackage{snapshot}
\newcommand{\mystylename}{icml2016}
\renewcommand{\mystylename}{icml2016-arxiv}

\documentclass[english]{article}
\usepackage[T1]{fontenc}
\usepackage[latin9]{inputenc}
\usepackage{color}
\usepackage{array}
\usepackage{url}
\usepackage{amstext}
\usepackage{graphicx}
\usepackage[authoryear]{natbib}

\usepackage{pdfpages}

\makeatletter

\providecommand{\tabularnewline}{\\}

\usepackage{times}
\usepackage{graphicx} 
\usepackage{subfig}

\usepackage{natbib}

\usepackage{times}
\usepackage{epsfig}
\usepackage{graphicx}
\usepackage{amsmath}
\usepackage{amssymb}

\usepackage[nohyperref,accepted]{\mystylename}

\usepackage{multirow}
\usepackage{algpseudocode}
\usepackage{algorithm}
\usepackage{lipsum}
\usepackage{hhline}
\usepackage{paralist}
\usepackage{array}
\usepackage{placeins}

\makeatother

\usepackage{babel}

\newcommand{\maintitle}{Augmenting Supervised Neural Networks with Unsupervised Objectives \\for Large-scale Image Classification} 

\newcommand{\mytitle}{\maintitle}

\begin{document}
\twocolumn[
\icmltitlerunning{\renewcommand{\\}{} \maintitle}
\icmltitle{\mytitle}
\vspace{-1em}
\icmlauthor{Yuting Zhang}{yutingzh@umich.edu}
\icmlauthor{Kibok Lee}{kibok@umich.edu}
\icmlauthor{Honglak Lee}{honglak@eecs.umich.edu}
\icmladdress{Department of Electrical Engineering and Computer Science, University of Michigan, Ann Arbor, MI, USA}
\vspace{-1em}
\vskip 0.3in

\title{\renewcommand{\\}{} \mytitle}
 ]

\newcommand{\supp}[1]{the supplementary materials}
\newcommand{\suppx}[2]{the supplementary materials}
\newcommand{\suppd}[1]{}

\newcommand{\cutsectionbefore}{}
\newcommand{\cutsectionafter}{}
\newcommand{\cutitem}{}
\newcommand{\cutfigurebelow}{}
\newcommand{\cutcaptionabove}{}
\newcommand{\cuteq}{}

\renewcommand{\supp}[1]{Appendix~\ref{#1}}
\renewcommand{\suppx}[2]{Appendix~\ref{#1} (#2)}
\renewcommand{\suppd}[1]{ (Appendix~\ref{#1})}

\renewcommand{\cutsectionbefore}{}
\renewcommand{\cutsectionafter}{}
\renewcommand{\cutitem}{}
\renewcommand{\cutfigurebelow}{}
\renewcommand{\cutcaptionabove}{}
\renewcommand{\cuteq}{}

\cfoot{\vspace{2em}\thepage}
 
\begin{abstract}
Unsupervised learning and supervised learning are key research topics in deep learning. 
However, as high-capacity supervised neural networks trained with a large amount of labels have achieved remarkable success in many computer vision tasks, the availability of large-scale labeled images reduced the significance of unsupervised learning.
Inspired by the recent trend toward revisiting the importance of unsupervised learning, we investigate joint supervised and unsupervised learning in a large-scale setting by augmenting existing neural networks with decoding pathways for reconstruction.
First, we demonstrate that the intermediate activations of pretrained large-scale classification networks preserve almost all the information of input images except a portion of local spatial details.
Then, by end-to-end training of the entire augmented architecture with the reconstructive objective, we show improvement of the network performance for supervised tasks.
We evaluate several variants of autoencoders, including the recently proposed ``what-where" autoencoder that uses the encoder pooling switches, to study the importance of the architecture design.
Taking the 16-layer VGGNet trained under the ImageNet ILSVRC 2012 protocol as a strong baseline for image classification, our methods improve the validation-set accuracy by a noticeable margin.
\end{abstract}
 
\cutsectionbefore
\section{Introduction}
\cutsectionafter

Unsupervised and supervised learning have been two associated key topics in deep learning.  
One important application of deep unsupervised learning over the past decade was to pretrain a deep neural network, which was then finetuned with supervised tasks (such as classification).
Many deep unsupervised models were proposed, such as stacked (denoising) autoencoders~\cite{SAE,SDAE}, deep belief networks~\citep{hinton2006fast,cdbn}, sparse encoder-decoders~\cite{unsup-invariant,fast-infer-sparse}, and deep Boltzmann machines~\citep{dbm}.
These approaches significantly improved the performance of neural networks on supervised tasks when the amount of available labels were not large.

However, over the past few years, supervised learning without any unsupervised pretraining has achieved even better performance, and it has become the dominating approach to train deep neural networks for real-world tasks, such as image classification \citep{alexnet} and object detection \citep{rcnn-pami}. 
Purely supervised learning allowed more flexibility of network architectures, e.g., the inception unit \citep{googlenet} and the residual structure \citep{residual-net}, which were not limited by the modeling assumptions of unsupervised methods. 
Furthermore, the recently developed batch normalization (BN) method \citep{bn}  has made the neural network learning further easier. 
As a result, the once popular framework of unsupervised pretraining has become less significant and even overshadowed \citep{dl-nature} in the field.

Several attempts (e.g., \citet{layerwise-semi,larochelle2008classification,icml2013pgbm,multi-pred-dbm}) had been made to couple the unsupervised and supervised learning in the same phase, making unsupervised objectives able to impact the network training after supervised learning took place. 
These methods  unleashed new potential of unsupervised learning, but
they have not yet been shown to scale to large amounts of labeled and unlabeled data.
\citet{ladder-semi} recently proposed an architecture that is easy to couple with a classification network by extending the stacked denoising autoencoder with lateral connections, i.e., from encoder to the same stages of the decoder, and their methods showed promising semi-supervised learning results. 
Nonetheless, the existing validations \citep{ladder-semi,ladder-deconstruct} were mostly on small-scale datasets like MNIST. 
Recently, \citet{what-where} proposed the ``what-where'' autoencoder (SWWAE) by extending the stacked convolutional autoencoder using \citet{ada-deconv}'s ``unpooling'' operator, which recovers the locational details (which was lost due to max-pooling) using the pooling switches from the encoder. 
While achieving promising results on the CIFAR dataset with extended unlabeled data \citep{tiny80}, SWWAE has not been demonstrated effective for larger-scale supervised tasks. 

In this paper, inspired by the recent trend toward simultaneous supervised and unsupervised neural network learning, we augment challenge-winning neural networks with decoding pathways for reconstruction, demonstrating the feasibility of improving high-capacity networks for large-scale image classification. 
Specifically, we take a segment of the classification network as the encoder and use the mirrored architecture as the decoding pathway to build several autoencoder variants. 
The autoencoder framework is easy to construct by augmenting an existing network without involving complicated components. 
Decoding pathways can be trained either separately from or together with the encoding/classification pathway by the standard stochastic gradient descent methods without special tricks, such as noise injection and activation normalization. 

This paper first investigates reconstruction properties of the large-scale deep neural networks. 
Inspired by \citet{invert-cnn}, we use the auxiliary decoding pathway of the stacked autoencoder to reconstruct images from intermediate activations of the pretrained classification network. 
Using SWWAE, we demonstrate better image reconstruction qualities compared to the autoencoder using the unpooling operators with \emph{fixed} switches, which upsamples an activation to a fixed location within the kernel. 
This result suggests that the intermediate (even high-level) feature representations preserve nearly all the information of the input images except for the locational details ``neutralized'' by max-pooling layers. 

Based on the above observations, we further improve the quality of reconstruction, an indication of the mutual information between the input and the feature representations \citep{SDAE}, by finetuning the \emph{entire} augmented architecture with supervised and unsupervised objectives. In this setting, the image reconstruction loss can also impact the classification pathway. 
To the contrary of conventional beliefs in the field, we demonstrate that the unsupervised learning objective posed by the auxiliary autoencoder is an effective way to help the classification network obtain better local optimal solutions for supervised tasks. 
To the best of our knowledge, this work is the first to show that unsupervised objective can improve the image classification accuracy of deep convolutional neural networks on large-scale datasets, such as ImageNet \citep{imagenet}. We summarize our main contributions as follows:
\begin{itemize} \cutitem
\item We show that the feature representations learned by high-capacity neural networks preserve the input information extremely well, despite the spatial invariance induced by pooling. 
Our models can perform high-quality image reconstruction (i.e., ``inversion'') from intermediate activations with the unpooling operator using the known switches from the encoder. \item We successfully improve the large-scale image classification performance of a state-of-the-art classification network by finetuning the augmented network with a reconstructive decoding pathway to make its intermediate activations preserve the input information better.  \item We study several variants of the resultant autoencoder architecture, including instances of SWWAE and more basic versions of autoencoders, and provide insight on the importance of the pooling switches and the layer-wise reconstruction loss. 
\end{itemize}

\section{Related work}
\cutsectionafter

In terms of using image reconstruction to improve classification, our work is related to supervised sparse coding and dictionary learning work, which is known to extract sparse local features from image patches by sparsity-constrained reconstruction loss functions. 
The extracted sparse features are then used for classification purposes.
\citet{sup-dict} proposed to combine the reconstruction loss of sparse coding and the classification loss of sparse features in a unified objective function. \citet{sup-trans-invariant-sparse} extended this supervised sparse coding with max-pooling to obtain translation-invariant local features.

\citet{deconv-net} proposed deconvolutional networks for unsupervised feature learning that consist of multiple layers of convolutional sparse coding with max-pooling. 
Each layer is trained to reconstruct the output of the previous layer. \citet{ada-deconv} further introduced the \textquotedblleft unpooling with switches\textquotedblright{} layer to deconvolutional networks to enable end-to-end training. 

As an alternative to sparse coding and discriminative convolutional networks, autoencoders \citep{deep-ai} are another class of models for representation learning, in particular for the non-linear principal component analysis \citep{npca,hierarchical-npca} by minimizing the reconstruction errors of a bottlenecked neural network.
The stacked autoencoder (SAE) \citep{SAE} is amenable for hierarchical representation learning. With pooling-induced sparsity bottlenecks \citep{winner-take-all-ae}, the convolutional SAE \citep{scae} can learn features from middle-size images. 
In these unsupervised feature learning studies, sparsity is the key regularizer to induce meaningful features in a hierarchy. 

By injecting noises or corruptions to the input, denoising autoencoders \citep{DAE,SDAE} can learn robust filters to recover the uncorrupted input. \citet{ladder-net} further added noises to intermediate layers of denoising auto-encoders with lateral connections, which was called \textquotedblleft ladder network\textquotedblright .
\citet{ladder-semi} combined a classification task with the ladder network for semi-supervised learning, and they showed improved classification accuracy on MNIST and CIFAR-10.  
Here, supervision from the labeled data is the critical objective that prevents the autoencoder from learning trivial features.

\citet{what-where} proposed the SWWAE, a convolutional autoencoder with unpooling layer, and combined it with classification objective for semi-supervised learning.
This model integrates a discriminative convolutional network (for classification) and a deconvolutional network (for reconstruction) and can be regarded as a unification of deconvolutional networks, autoencoders and discriminative convolutional networks. 
They demonstrated promising results on small scale datasets such as MNIST, SVHN and STL10. 

Improving representation learning with auxiliary tasks is not new~\citep{rule-injection}. The idea behind is that the harder the tasks are, the better representations a network can learn. 
As an alternative to the autoencoder, \citet{dsn}'s ``deeply supervised network'' incorporated classification objectives for intermediate layers, was able to improve the top-layer classification accuracy for reasonably large-scale networks \citep{dsn-vgg}. 
In earlier work, \citet{layerwise-semi} conducted layer-wise training by both classification and reconstruction objectives. 
Recently, more task-specific unsupervised objectives for image and video representation learning were developed by using spatial context \citep{unsup-by-context} and video continuity \citep{unsup-by-video}. In contrast, autoencoder-based methods are applicable in more general scenarios.

\cutsectionbefore
\section{Methods \label{sec:methods}}
\cutsectionafter

In this section, we describe the training objectives and architectures
of the proposed augmented network. In Section~\ref{sub:general-obj},
we briefly review the architectures of recent networks for vision
tasks, and present the general form of our method. In Section~\ref{sub:arch},
we augment the classification network with auxiliary pathways composed
of deconvolutional architectures to build fully mirrored autoencoders,
on which we specify the auxiliary objective functions. 

\begin{figure}
\begin{centering}
\includegraphics[width=1\columnwidth]{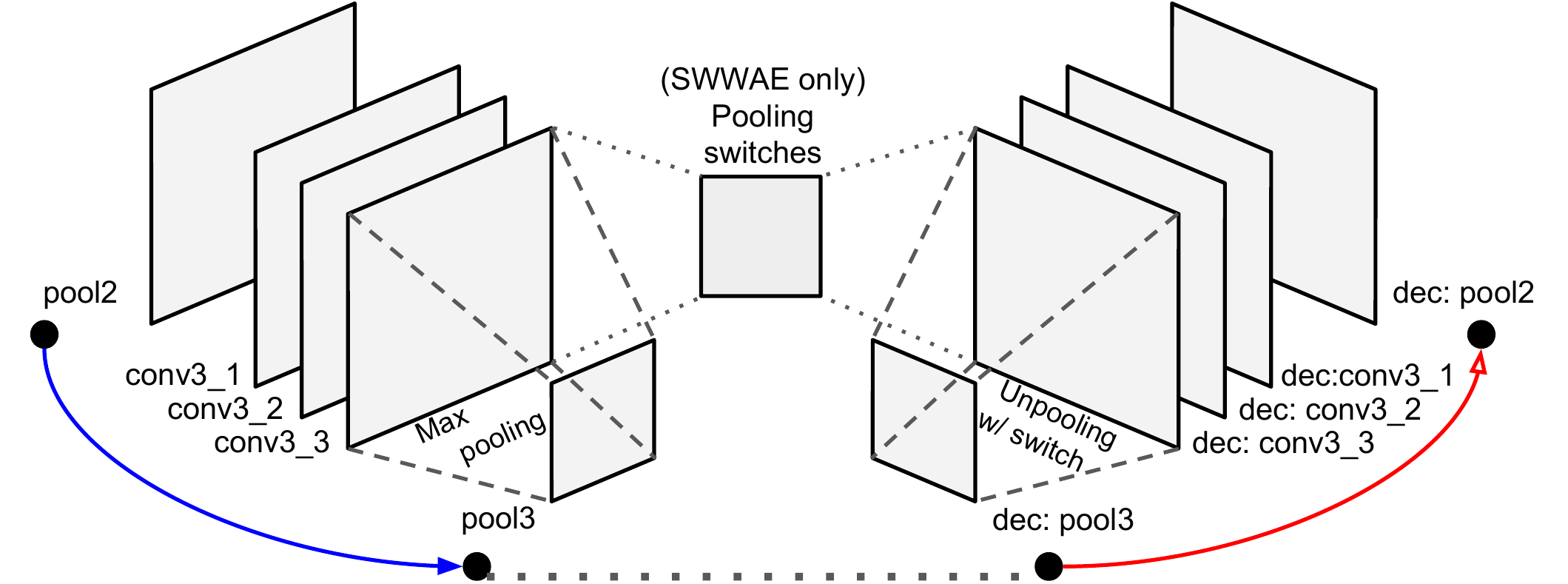}
\par\end{centering}

\cutcaptionabove
\caption{\label{fig:micro-layers}Example micro-architectures in macro-layers
(the 3\protect\textsuperscript{rd} macro-layer of VGGNet and its
mirrored decoder). \emph{Encoder:} a number of convolutional layers
followed by a max-pooling layer. \emph{Decoder:} the same number of
deconvolutional layers preceded by an unpooling layer, where the known
pooling switches given by the associated pooling layer are used for
SWWAE. }
\cutfigurebelow
\vspace*{0.1in}
\end{figure}

\cutsectionbefore
\subsection{Unsupervised loss for intermediate representations \label{sub:general-obj}}
\vspace*{-0.15in}
\cutsectionafter

Deep neural networks trained with full supervision achieved the state-of-the-art
image classification performance. Commonly used network architectures
\citep{alexnet} contain a single pathway of convolutional layers
succeeded by nonlinear activation functions and interleaved with max-pooling layers to gradually transform features into high-level representations
and gain spatial invariance at different scales. Recent networks \citep{vggnet,googlenet,residual-net,rethink-inception}
often nest a group of convolutional layers before applying a max-pooling
layer. As these layers work together as the feature extractor for a particular
scale, we refer to the group as a \emph{macro}-layer (see the left
half of Figure~\ref{fig:micro-layers}). Fully-connected inner-product
layer and/or global average-pooling layer follow the convolution-pooling
macro-layers to feed the top-layer classifier. A network of $L$
convolution-pooling macro-layers is defined as 
\begin{equation}
a_{l}=f_{l}(a_{l-1};\phi_{l}),\;\mbox{for }l=1,2,\ldots,L+1,\label{eq:cnn}
\end{equation}
where $a_{0}=x$ is the input, $f_{l}(l=1,2,\ldots,L)$ with the parameter
$\phi_{l}$ is the $l$\textsuperscript{th} macro-layer, and $f_{L+1}$
denotes the rest of the network, including the inner-product and classification
layers. The classification loss is $C(x,y)=\ell(a_{L+1},y)$, where
$y$ is the ground truth label, and $\ell$ is the cross-entropy loss
when using a softmax classifier. 

Let $x_{1},x_{2},\ldots,x_{N}$ denote a set of training images associated
with categorical labels $y_{1},y_{2},\ldots,y_{N}$. The neural network
is trained by minimizing $\frac{1}{N}\sum_{i=1}^{N}C(x_{i},y_{i})$,
where we omit the L2-regularization term on the parameters. Though
this objective can effectively learn a large-scale network by gradient
descent with a huge amount of labeled data, it has two limitations.
On the one hand, the training of lower intermediate layers might be
problematic, because the gradient signals from the top layer can become
vanished \citep{gradient-vanish} on its way to the bottom layer.
Regularization by normalization \citep{bn} can alleviate this problem,
but will also lead to large yet noisy gradients when networks are
deep \citep{residual-net}. On the other hand, the data space is informative
by itself, but the fully supervised objective guides the representation
learning purely by the labels. 

A solution to both problems is to incorporate auxiliary unsupervised
training objectives to the intermediate layers. More specifically,
 the objective function becomes 
 \cuteq
\begin{equation}
\frac{1}{N}\sum_{i=1}^{N}\left(C(x_{i},y_{i})+\lambda U(x_{i})\right),\label{eq:semi-obj}
\end{equation}
\cuteq
where $U(\cdot)$ is the unsupervised objective function associating
with one or more auxiliary pathways that are attached to the convolution-pooling
macro-layers in the original classification network.  

\begin{figure}[t]
\begin{centering}
\includegraphics[width=1\columnwidth]{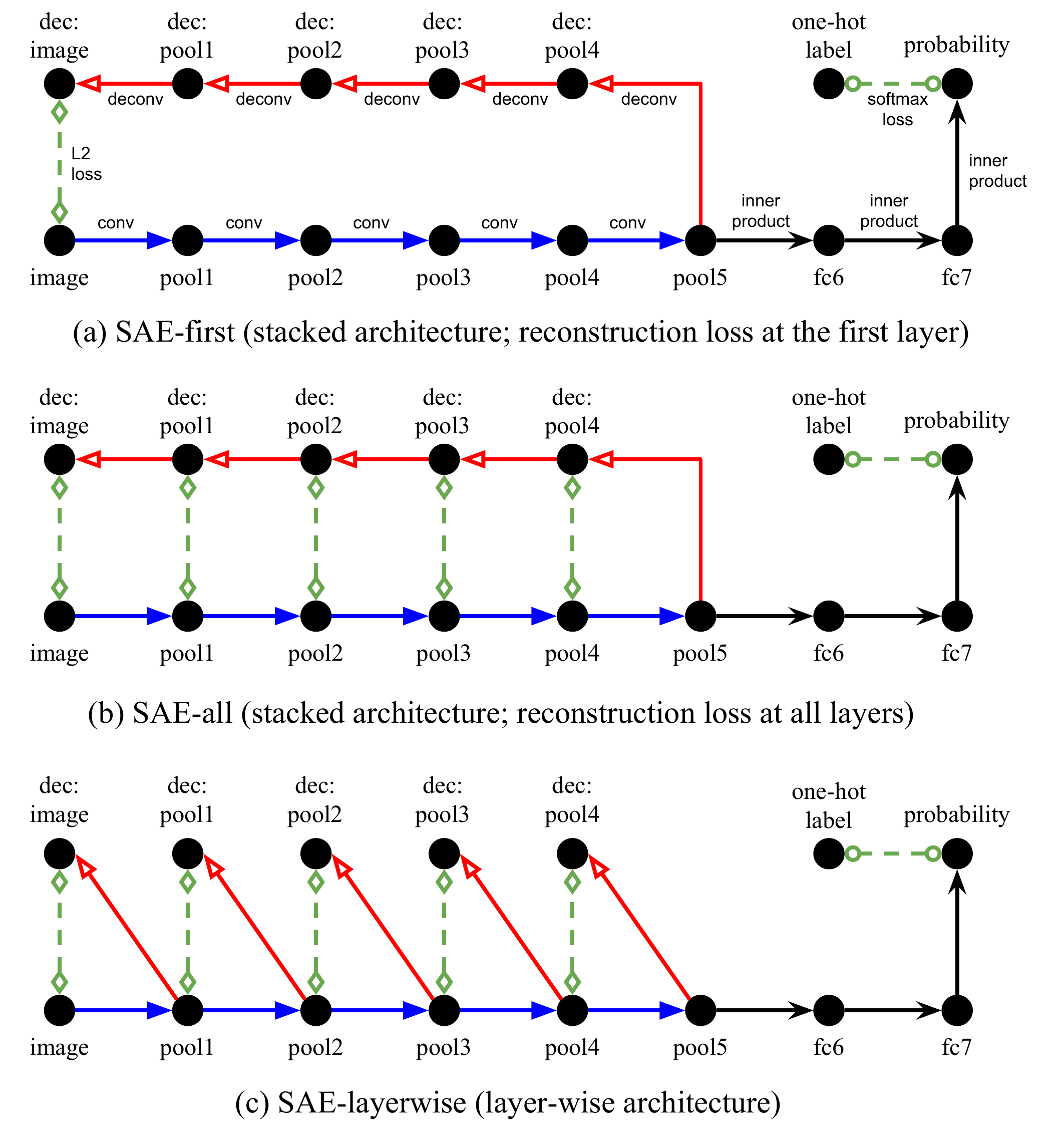}
\par\end{centering}

\cutcaptionabove
\vspace*{-0.05in}
\caption{\label{fig:swae}Model architectures of networks augmented with autoencoders.
\protect\includegraphics[bb=0bp 5bp 23bp 23bp,height=0.8em]{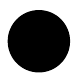}~:
nodes; \protect\includegraphics[bb=0bp 5bp 41bp 23bp,height=0.8em]{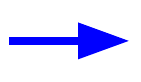}~:
encoder macro-layer; \protect\includegraphics[bb=0bp 5bp 36bp 24bp,height=0.8em]{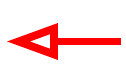}~:
decoder macro-layer; \protect\includegraphics[bb=0bp 5bp 36bp 24bp,height=0.8em]{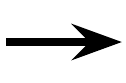}~:
inner-product layer; \protect\includegraphics[bb=0bp 5bp 52bp 23bp,height=0.8em]{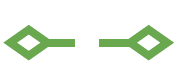}~:
reconstruction loss; \protect\includegraphics[bb=0bp 5bp 44bp 23bp,height=0.8em]{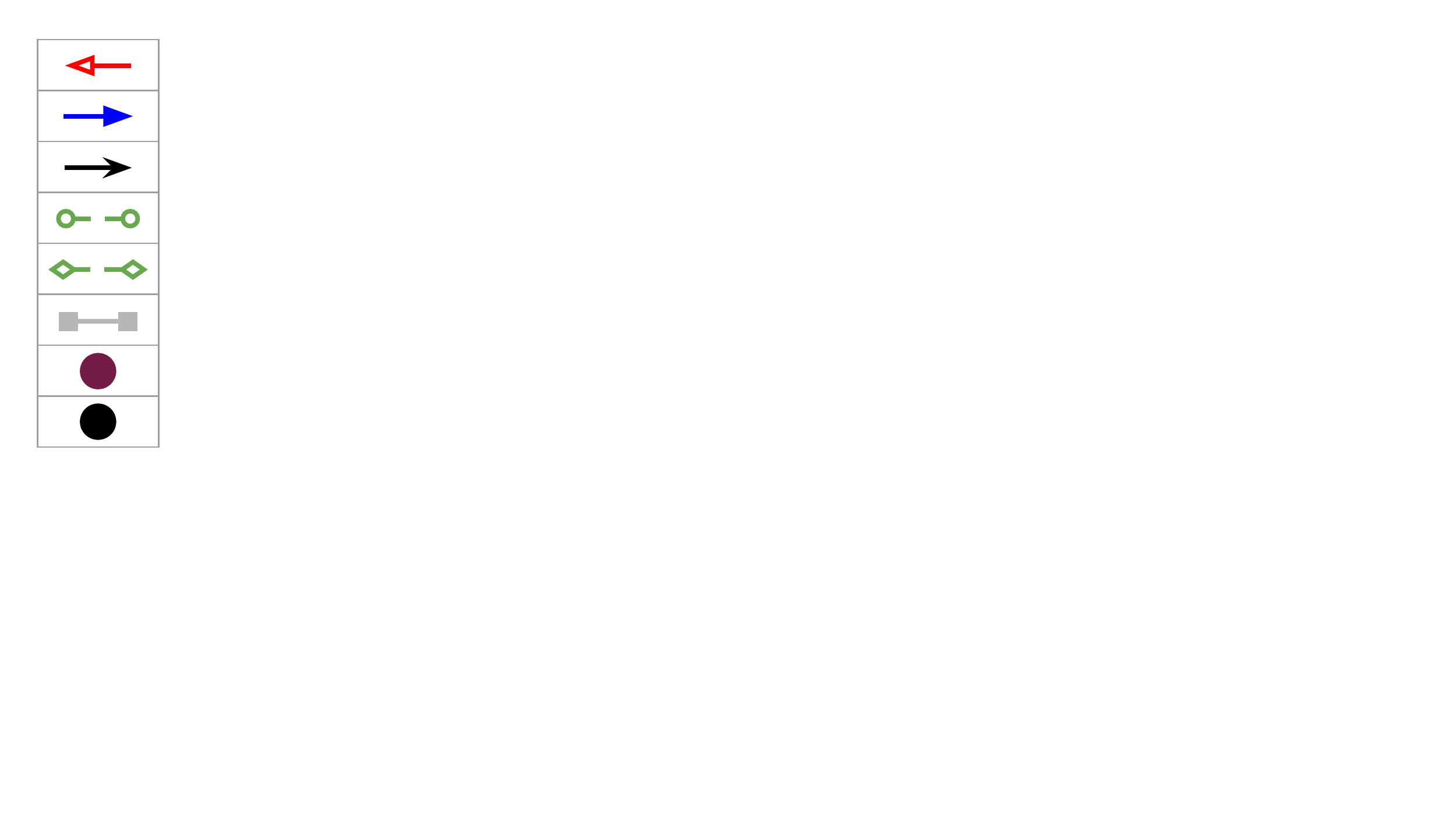}~:
classification loss.}
\cutfigurebelow
\end{figure}

\cutsectionbefore
\subsection{Network augmentation with autoencoders \label{sub:arch}}
\cutsectionafter

Given the network architecture for classification defined in Eq.~(\ref{eq:cnn}),
we take the sub-network composed of all the convolution-pooling macro-layers
as the encoding pathway, and generate a fully mirrored decoder network
as an auxiliary pathway of the original network. The inner-product
layers close to the top-level classifier may be excluded from the autoencoder,
since they are supposed to be more task-relevant. 

Taking a network of five macro-layers as an example (e.g., VGGNet),
Figure~\ref{fig:swae}a shows the network augmented with a stacked
autoencoder. The decoding starts from the pooled feature map from
the $5$\textsuperscript{th} macro-layer (pool5) all the way down
to the image input. Reconstruction errors are measured at the network
input (i.e., the first layer) so that we term the model as ``\emph{SAE-first}''.
More specifically, the decoding pathway is 
\cuteq
\begin{equation}
\hat{a}_{L}=a_{L},\hat{a}_{l-1}=f_{l}^{dec}(\hat{a}_{l};\psi_{l}),\hat{x}=\hat{a}_{0}.\label{eq:decoder-stacked}
\end{equation}
with the loss $U_{\textrm{SAE-first}}(x)= \Vert\hat{x}-x\Vert_{2}^{2}$. Here, $\psi_{l}$'s are decoder parameters.

\begin{figure}[tb]
\begin{centering}
\includegraphics[width=0.96\columnwidth]{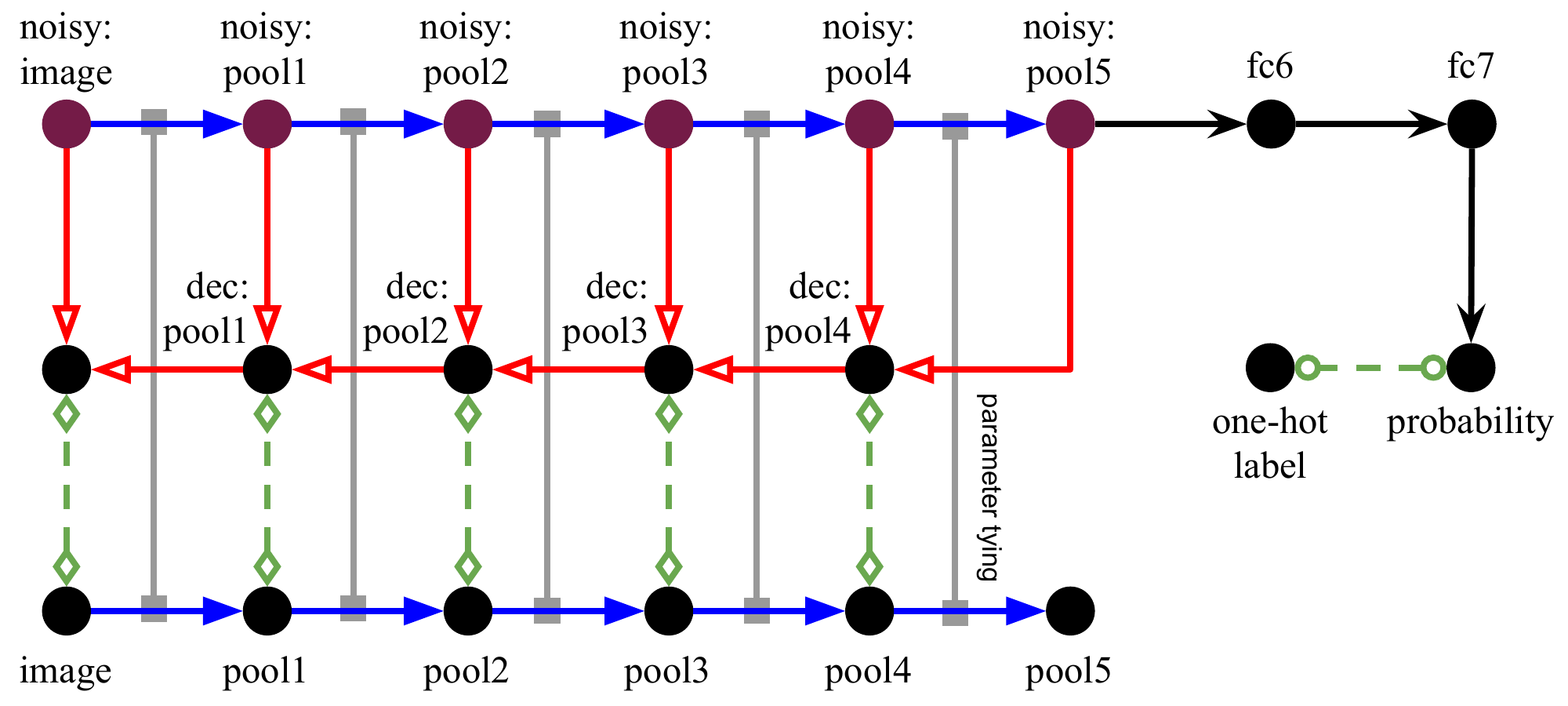}
\par\end{centering}

\cutcaptionabove
\caption{\label{fig:ladder}Ladder network architectures \citet{ladder-semi}.
\protect\includegraphics[bb=0bp 5bp 23bp 23bp,height=0.8em]{figures/swae-legend/node}~:
nodes; \protect\includegraphics[bb=0bp 5bp 23bp 23bp,height=0.8em]{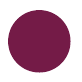}~:
noisy nodes; \protect\includegraphics[bb=0bp 5bp 41bp 23bp,height=0.8em]{figures/swae-legend/encode}~:
encoder macro-layer; \protect\includegraphics[bb=0bp 5bp 36bp 24bp,height=0.8em]{figures/swae-legend/decode}~:
decoder macro-layer; \protect\includegraphics[bb=0bp 5bp 36bp 24bp,height=0.8em]{figures/swae-legend/inner-prod}~:
inner-product layer; \protect\includegraphics[bb=0bp 5bp 52bp 23bp,height=0.8em]{figures/swae-legend/recon-loss}~:
reconstruction loss; \protect\includegraphics[bb=0bp 5bp 44bp 23bp,height=0.8em]{figures/swae-legend/cls-loss}~:
classification loss; \protect\includegraphics[bb=0bp 5bp 46bp 23bp,height=0.8em]{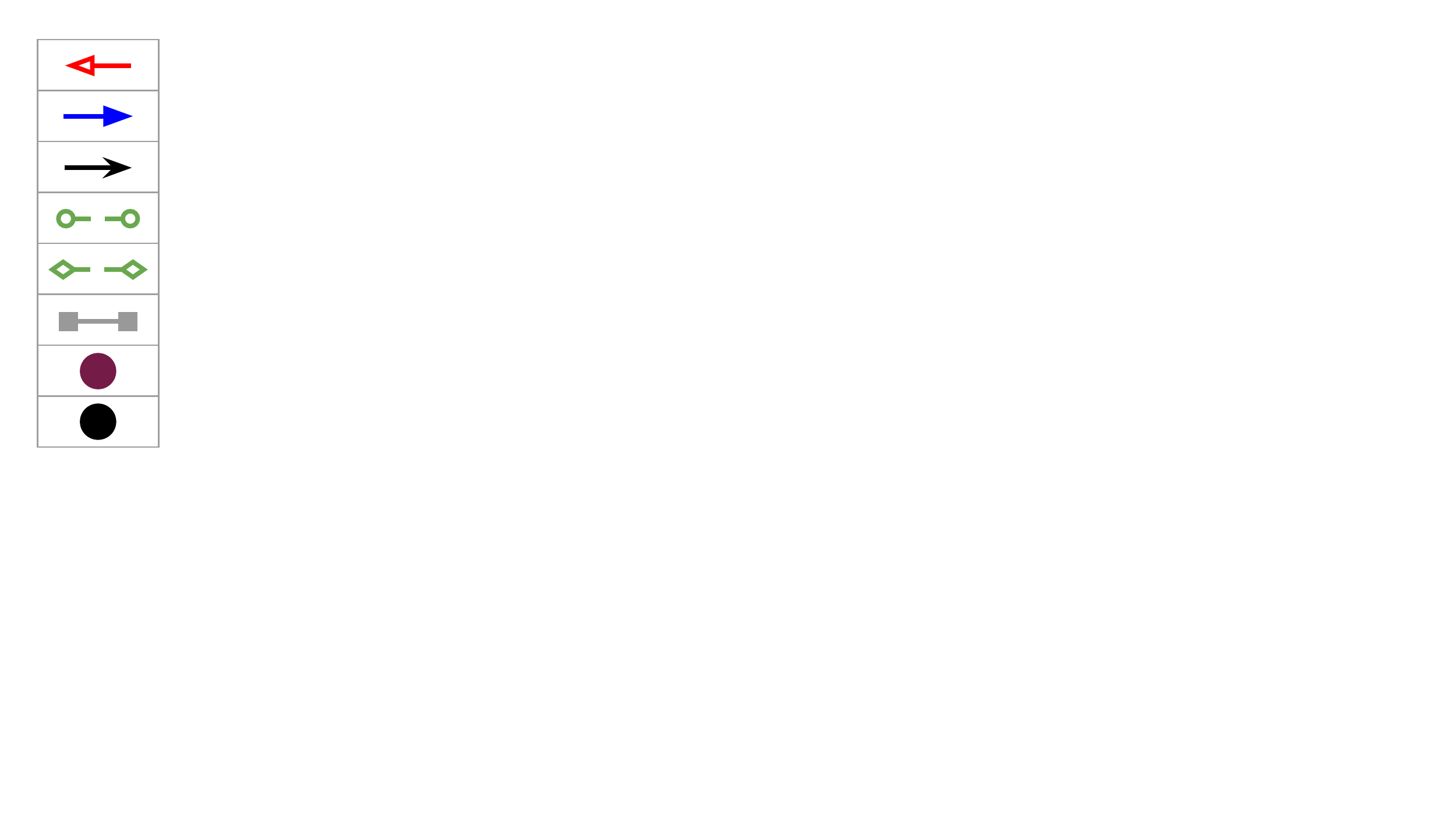}~:
parameter tying.}
\cutfigurebelow
\end{figure}

The auxiliary training signals of SAE-first emerge from the bottom
of the decoding pathway, and they get merged with the top-down signals
for classification at the last convolution-pooling macro-layer into
the encoder pathway. To allow more gradient to flow directly into the
preceding macro-layers, we propose the ``\emph{SAE-all}'' model
by replacing the unsupervised loss by $U_{\textrm{SAE-all}}(x)=\sum_{l=0}^{L-1}\gamma_{l}\Vert\hat{a}_{l}-a_{l}\Vert_{2}^{2}$
, which makes the autoencoder have an even better mirrored architecture
by matching activations for all the macro-layer (illustrated in Figure~\ref{fig:swae}b). 

In Figure~\ref{fig:swae}c, we propose one more autoencoder variant
with layer-wise decoding architecture, termed ``\emph{SAE-layerwise}''.
It reconstructs the output activations of every macro-layer to its
input.  The auxiliary loss of SAE-layerwise is the same as SAE-all, i.e.,
$U_{\textrm{SAE-layerwise}}(x)=U_{\textrm{SAE-all}}(x)$, but the decoding
pathway is replaced by $\hat{a}_{l-1}=f_{l}^{dec}(a_{l};\psi_{l})$. 

SAE-first/all encourages top-level convolution features to preserve as much information as possible. 
In contrast, the auxiliary pathways in SAE-layerwise focus on inverting the clean intermediate activations (from the encoder) to the input of the associated macro-layer, admitting parallel layer-wise training. 
We investigated both in Section~\ref{sec:cls} and take SAE-layerwise decoders as architectures for efficient pretraining.

In Figure~\ref{fig:micro-layers}, we illustrate the detailed architecture of $f_{3}(\cdot)$ and $f_{3}^{dec}(\cdot)$ for \citet{vggnet}'s 16-layer VGGNet. 
Inspired by \citet{ada-deconv}, we use \citet{what-where}'s SWWAE as the default for the micro-architecture. 
More specifically, we record the pooling switches (i.e., the locations of the local maxima) in the encoder, and unpool activations by putting the elements at the recorded locations and filling the blanks with zeros. 
Unpooling with known switches can recover the local spatial variance eliminated by the max-pooling layer, avoiding the auxiliary objectives from deteriorating the spatial invariance of the encoder filters, which is arguably important for classification. 
We studied the autoencoders with fixed and known unpooling switch, respectively. 
In Section~\ref{sec:recon} we efficiently trained the autoencoders augmented from a pretrained deep non-BN network, where the decoder is hard to learn from scratch.

\citet{ladder-semi}'s ladder network (Figure \ref{fig:ladder}) is a more sophisticated way to augment existing sequential architectures with autoencoders. 
It is featured by the lateral connections (vertical in Figure \ref{fig:ladder}) and the combinator functions that merge the lateral and top-down activations. 
Due to the lateral connections, noise must be added to the encoder; 
otherwise, the combinator function can trivially copy the clean activations from the encoder. 
In contrast, no autoencoder variant used in our work has ``lateral" connections, which makes the overall architectures of
our models simpler and more standard. 
In SWWAE, the pooling switch connections do not bring the encoder input directly to the decoder, so they cannot be taken as the lateral connections like in the ``ladder network''. 
Moreover, noise injection is also unnecessary for our models. 
We leave it as an open question whether denoising objectives can help with the augmented (what-where) autoencoder for large-scale data.

\cutsectionbefore
\section{Experiments \label{sec:exp}}
\cutsectionafter

In this section, we evaluated different variants of the augmented network for image reconstruction and classification on ImageNet ILSVRC 2012 dataset, using the training set for training, and validation set for evaluation. 
Our experiments were mainly based on the 16-layer VGGNet \citep{vggnet}.\footnote{The pretrained network was obtained from \url{http://www.robots.ox.ac.uk/~vgg/research/very_deep/}.} 
To compare with existing methods on inverting neural networks \cite{invert-cnn}, we also partially used  \citet{alexnet}'s network, termed AlexNet, trained on ILSVRC2012 training set. 
Our code and trained models can be obtained at \url{http://www.ytzhang.net/software/recon-dec/}

\cutsectionbefore
\subsection{Training procedure \label{sec:train}}
\cutsectionafter

Training a deep neural network is non-trivial. 
Therefore, we propose the following strategy to make the networks augmented from the classification network efficiently trainable. 

\begin{enumerate} \cutitem
\item \label{enu:init}We initialized the encoding pathway with the pretrained classification network, and the decoding pathways with Gaussian random initialization. \cutitem
\item \label{enu:ael-dec}For any variant of the augmented network, we fixed the parameters for the classification pathway and trained the layer-wise decoding pathways of the SAE-layerwise network. \cutitem
\item \label{enu:sae}For SAE-first/all, we initialized the decoding pathway with the pretrained SAE-layerwise parameters and finetuned the decoder. (Skip this step for SAE-layerwise.) \cutitem
\item \label{enu:ft}We finetuned all the decoding and the encoding/classification pathways together with a reduced learning rate. \cutitem
\end{enumerate}
Up to Step~\ref{enu:sae}, we trained the decoding pathways with the classification pathway fixed.
For all the four steps, we trained the networks by mini-batch stochastic gradient descent (SGD) with the momentum $0.9$.

\begin{figure*}[!t]

{ \small{} 

\begin{centering}
\begin{tabular}{c|ccccccccc}
\hline 
Layer & image & pool1 & pool2 & conv3 & conv4 & pool5 & fc6 & fc7 & fc8\tabularnewline
\hline 
\hspace{-5pt}\begin{minipage}[t]{6.5em}\begin{center}
\vspace{1.2em}
\citet{invert-cnn}
\par\end{center}\end{minipage}\hspace{-5pt} & \hspace{-5pt}\begin{minipage}[t]{0.09\textwidth}
\noindent \begin{center}
\vspace{-1em}
\includegraphics[width=1\columnwidth]{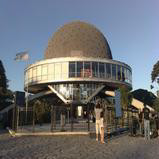}\vspace{2pt}
\par\end{center}

\end{minipage}\hspace{-5pt} & \hspace{-5pt}\begin{minipage}[t]{0.09\textwidth}
\noindent \begin{center}
\vspace{-1em}
\includegraphics[width=1\columnwidth]{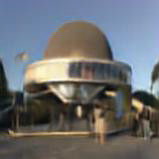}\vspace{2pt}
\par\end{center}

\end{minipage}\hspace{-5pt} & \hspace{-5pt}\begin{minipage}[t]{0.09\textwidth}
\noindent \begin{center}
\vspace{-1em}
\includegraphics[width=1\columnwidth]{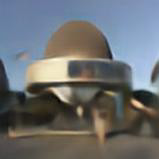}\vspace{2pt}
\par\end{center}

\end{minipage}\hspace{-5pt} & \hspace{-5pt}\begin{minipage}[t]{0.09\textwidth}
\noindent \begin{center}
\vspace{-1em}
\includegraphics[width=1\columnwidth]{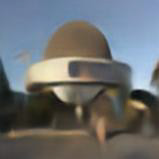}\vspace{2pt}
\par\end{center}

\end{minipage}\hspace{-5pt} & \hspace{-5pt}\begin{minipage}[t]{0.09\textwidth}
\noindent \begin{center}
\vspace{-1em}
\includegraphics[width=1\columnwidth]{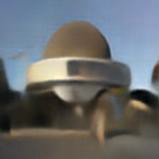}\vspace{2pt}
\par\end{center}

\end{minipage}\hspace{-5pt} & \hspace{-5pt}\begin{minipage}[t]{0.09\textwidth}
\noindent \begin{center}
\vspace{-1em}
\includegraphics[width=1\columnwidth]{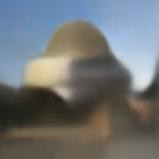}\vspace{2pt}
\par\end{center}

\end{minipage}\hspace{-5pt} & \hspace{-5pt}\begin{minipage}[t]{0.09\textwidth}
\noindent \begin{center}
\vspace{-1em}
\includegraphics[width=1\columnwidth]{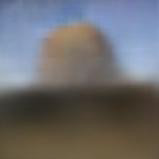}\vspace{2pt}
\par\end{center}

\end{minipage}\hspace{-5pt} & \hspace{-5pt}\begin{minipage}[t]{0.09\textwidth}
\noindent \begin{center}
\vspace{-1em}
\includegraphics[width=1\columnwidth]{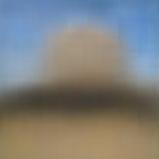}\vspace{2pt}
\par\end{center}

\end{minipage}\hspace{-5pt} & \hspace{-5pt}\begin{minipage}[t]{0.09\textwidth}
\noindent \begin{center}
\vspace{-1em}
\includegraphics[width=1\columnwidth]{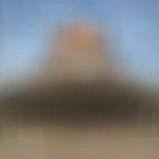}\vspace{2pt}
\par\end{center}

\end{minipage}\hspace{-5pt}\tabularnewline
\hline 
\hspace{-5pt}\begin{minipage}[t]{6.5em}\begin{center}
\vspace{-0.3em}
SWWAE-first 
(\textbf{known} unpooling switches)
\par\end{center}\end{minipage}  & \hspace{-5pt}\begin{minipage}[t]{0.09\textwidth}
\noindent \begin{center}
\vspace{-1em}
\includegraphics[width=1\columnwidth]{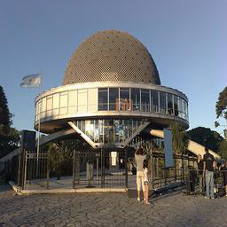}\vspace{2pt}
\par\end{center}

\end{minipage}\hspace{-5pt} & \hspace{-5pt}\begin{minipage}[t]{0.09\textwidth}
\noindent \begin{center}
\vspace{-1em}
\includegraphics[width=1\columnwidth]{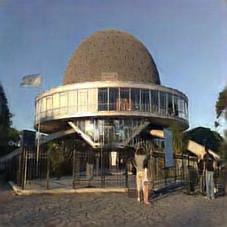}\vspace{2pt}
\par\end{center}

\end{minipage}\hspace{-5pt} & \hspace{-5pt}\begin{minipage}[t]{0.09\textwidth}
\noindent \begin{center}
\vspace{-1em}
\includegraphics[width=1\columnwidth]{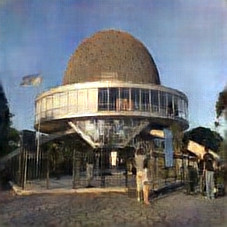}\vspace{2pt}
\par\end{center}

\end{minipage}\hspace{-5pt} & \hspace{-5pt}\begin{minipage}[t]{0.09\textwidth}
\noindent \begin{center}
\vspace{-1em}
\includegraphics[width=1\columnwidth]{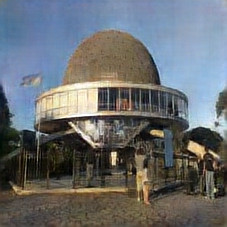}\vspace{2pt}
\par\end{center}

\end{minipage}\hspace{-5pt} & \hspace{-5pt}\begin{minipage}[t]{0.09\textwidth}
\noindent \begin{center}
\vspace{-1em}
\includegraphics[width=1\columnwidth]{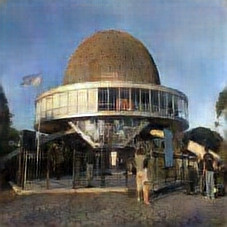}\vspace{2pt}
\par\end{center}

\end{minipage}\hspace{-5pt} & \hspace{-5pt}\begin{minipage}[t]{0.09\textwidth}
\noindent \begin{center}
\vspace{-1em}
\includegraphics[width=1\columnwidth]{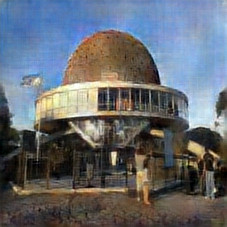}\vspace{2pt}
\par\end{center}

\end{minipage}\hspace{-5pt} & \hspace{-5pt}\begin{minipage}[t]{0.09\textwidth}
\noindent \begin{center}
\vspace{-1em}
\includegraphics[width=1\columnwidth]{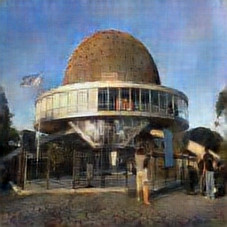}\vspace{2pt}
\par\end{center}

\end{minipage}\hspace{-5pt} & \hspace{-5pt}\begin{minipage}[t]{0.09\textwidth}
\noindent \begin{center}
\vspace{-1em}
\includegraphics[width=1\columnwidth]{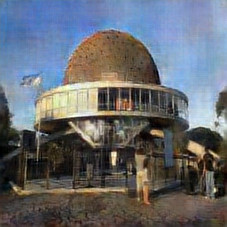}\vspace{2pt}
\par\end{center}

\end{minipage}\hspace{-5pt} & \hspace{-5pt}\begin{minipage}[t]{0.09\textwidth}
\noindent \begin{center}
\vspace{-1em}
\includegraphics[width=1\columnwidth]{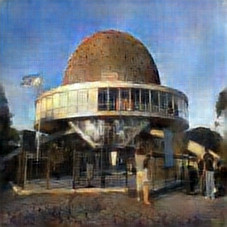}\vspace{2pt}
\par\end{center}

\end{minipage}\hspace{-5pt}\tabularnewline
\hline 
\end{tabular}
\par\end{centering}

 }
 
\cutcaptionabove
\caption{\label{fig:alex-recon1}AlexNet reconstruction on ImageNet ILSVRC2012 validation set. See \suppx{sub:more-image-recon}{Figure~\ref{fig:alex-recon2}} for more results.}
\end{figure*}

\begin{figure*}[!t]

{ \small{} 

\begin{centering}
\begin{tabular}{c|cccccc}
\hline 
Layer & image & pool1 & pool2 & pool3 & pool4 & pool5\tabularnewline
\hline 
\begin{minipage}[t]{8em}\begin{center}
\vspace{0em}
SAE-first
(\textbf{fixed} unpooling switches)
\par\end{center}\end{minipage} & \hspace{-5pt}\begin{minipage}[t]{0.12\textwidth}
\noindent \begin{center}
\vspace{-1em}
\includegraphics[width=1\columnwidth]{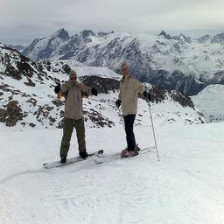}\vspace{2pt}

\par\end{center}
\end{minipage}\hspace{-5pt} & \hspace{-5pt}\begin{minipage}[t]{0.12\textwidth}
\noindent \begin{center}
\vspace{-1em}
\includegraphics[width=1\columnwidth]{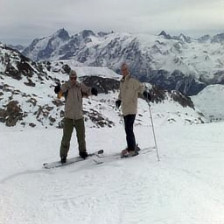}\vspace{2pt}

\par\end{center}
\end{minipage}\hspace{-5pt} & \hspace{-5pt}\begin{minipage}[t]{0.12\textwidth}
\noindent \begin{center}
\vspace{-1em}
\includegraphics[width=1\columnwidth]{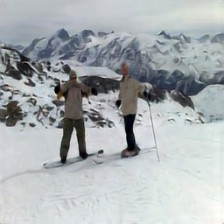}\vspace{2pt}

\par\end{center}
\end{minipage}\hspace{-5pt} & \hspace{-5pt}\begin{minipage}[t]{0.12\textwidth}
\noindent \begin{center}
\vspace{-1em}
\includegraphics[width=1\columnwidth]{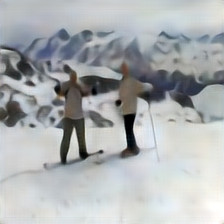}\vspace{2pt}

\par\end{center}
\end{minipage}\hspace{-5pt} & \hspace{-5pt}\begin{minipage}[t]{0.12\textwidth}
\noindent \begin{center}
\vspace{-1em}
\includegraphics[width=1\columnwidth]{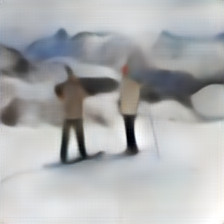}\vspace{2pt}

\par\end{center}
\end{minipage}\hspace{-5pt} & \hspace{-5pt}\begin{minipage}[t]{0.12\textwidth}
\noindent \begin{center}
\vspace{-1em}
\includegraphics[width=1\columnwidth]{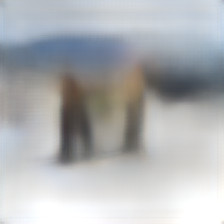}\vspace{2pt}

\par\end{center}
\end{minipage}\hspace{-5pt}\tabularnewline
\begin{minipage}[t]{8em}\begin{center}
\vspace{0em}
SWWAE-first
(\textbf{known} unpooling switches)
\par\end{center}\end{minipage} & \hspace{-5pt}\begin{minipage}[t]{0.12\textwidth}
\noindent \begin{center}
\vspace{-1em}
\includegraphics[width=1\columnwidth]{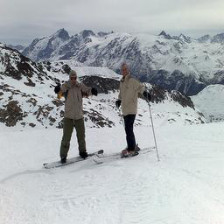}\vspace{2pt}

\par\end{center}
\end{minipage}\hspace{-5pt} & \hspace{-5pt}\begin{minipage}[t]{0.12\textwidth}
\noindent \begin{center}
\vspace{-1em}
\includegraphics[width=1\columnwidth]{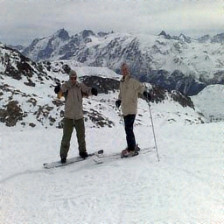}\vspace{2pt}

\par\end{center}
\end{minipage}\hspace{-5pt} & \hspace{-5pt}\begin{minipage}[t]{0.12\textwidth}
\noindent \begin{center}
\vspace{-1em}
\includegraphics[width=1\columnwidth]{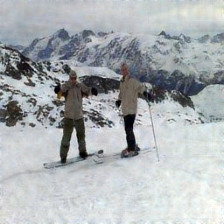}\vspace{2pt}

\par\end{center}
\end{minipage}\hspace{-5pt} & \hspace{-5pt}\begin{minipage}[t]{0.12\textwidth}
\noindent \begin{center}
\vspace{-1em}
\includegraphics[width=1\columnwidth]{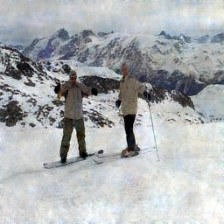}\vspace{2pt}

\par\end{center}
\end{minipage}\hspace{-5pt} & \hspace{-5pt}\begin{minipage}[t]{0.12\textwidth}
\noindent \begin{center}
\vspace{-1em}
\includegraphics[width=1\columnwidth]{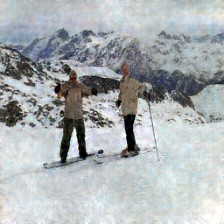}\vspace{2pt}

\par\end{center}
\end{minipage}\hspace{-5pt} & \hspace{-5pt}\begin{minipage}[t]{0.12\textwidth}
\noindent \begin{center}
\vspace{-1em}
\includegraphics[width=1\columnwidth]{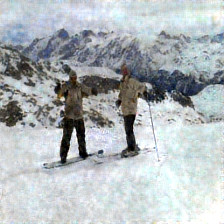}\vspace{2pt}

\par\end{center}
\end{minipage}\hspace{-5pt}\tabularnewline
\hline 
\end{tabular}
\par\end{centering}

 }
 
\cutcaptionabove
\caption{\label{fig:vgg-recon1}VGGNet reconstruction on ImageNet ILSVRC2012 validation set. See \suppx{sub:more-image-recon}{Figure~\ref{fig:alex-recon2}} for more results.}
\cutfigurebelow
\end{figure*}

In Step~\ref{enu:ael-dec}, the SAE-layerwise model has separate sub-pathways for decoding, so the training can be done in parallel for every macro-layer.
The decoding sub-network for each macro-layer was relatively ``shallow'' so that it is easy to learn. 
We found the learning rate annealing not critical for SAE-layerwise pretraining. 
Proper base learning rates could make it sufficiently converged within $1$ epoch. 
The chosen layer-wise learning rates VGGNet were summarized in \suppx{sec:vgg-param}{Table~\ref{tab:vgg-param}}. 
We used a small mini-batch size of $16$ for SGD. 

For very deep networks, training the decoding pathways of SAE-first/all from random initialization is difficult when batch normalization is absent (e.g., in the VGGNet). 
Initializing with SAE-layerwise as in Step~\ref{enu:sae} is critical to efficiently train the stacked decoding pathways of SAE-first and SAE-all. 

For SAE-all (Step~\ref{enu:sae}, \ref{enu:ft}) and SAE-layerwise (Step~\ref{enu:ft}), we balanced the reconstruction loss among different macro-layer, where the criterion was to make the weighted loss for every layer comparable to each other.
We summarized the balancing weights for VGGNet in \suppx{sec:vgg-param}{Table~\ref{tab:vgg-param}}. 
The SGD mini-batch size was set to a larger value (here, $64$) in Step~\ref{enu:ft} for better stability.

We adopted commonly used data augmentation schemes. 
As to VGGNet, we randomly resized the image to $[256,512]$ pixels with respect to the shorter edge, and then randomly cropped a $224\times224$ patch (or its horizontally mirrored image) to feed into the network.
As to AlexNet, we followed \citet{alexnet}'s data augmentation scheme, cropping an image at the center to make it square with the shorter edge unchanged, resizing the square to $256\times256$, and randomly sampling a $227\times227$ patch or its horizontally mirrored counterpart to feed the network. 
We ignored the RGB color jittering so as to always take ground truth natural images as the reconstruction targets. 

Our implementation was based on the Caffe framework \citep{caffe}.

\cutsectionbefore
\subsection{Image reconstruction via decoding pathways\label{sec:recon}}
\cutsectionafter
Using reconstructive decoding pathways, we can visualize the learned hierarchical features by inverting a given classification network, which is a useful way to understand the learned representations. 
The idea of reconstructing the encoder input from its intermediate activations was first explored by \citet{invert-cnn}, in contrast to visualizing a single hidden node \citep{vis-cnn} and dreaming out images \citep{invert-understand}.
As the best existing method for inverting neural networks with no skip link,  it used unpooling with fixed switches to upsample the intermediate activation maps. 
This method demonstrated how much information the features produced by each layer could preserve for the input. As shown in Figure~\ref{fig:alex-recon1} (the top row), not surprisingly, the details of the input image gradually diminished as the representations went through higher layers. 

The commonly used classification network mainly consists of convolution/inner-product and max-pooling operators. Based only on \citet{invert-cnn}'s visualization, it is hard to tell how much the two types of operators contribute to the diminishing of image details, respectively. 
Note that our SAE-first architecture is comparable to \citet{invert-cnn}'s model except for the better mirrored architectures between the encoder and decoder, which allow extending to SWWAE. 
Using the SWWAE-first network (``what-where'' version of SAE-first), we were able to revert the max-pooling more faithfully, and to study the amount of information that the convolutional filters and inner-product coefficients preserved. 

To compare with \citet{invert-cnn}, we augmented AlexNet to the corresponding SWWAE-first architecture.\footnote{The decoding pathway almost fully mirrored the classification network except the first layer (\texttt{conv1}). This convolutional layer used the stride $4$ rather than $1$, which approximates two additional $2 \times 2$ pooling layers. Therefore, we used three deconvolutional layers to inverse the \texttt{conv1} layer.} 
Unlike in Section~\ref{sec:methods}, we built SWWAE-first network starting from every layer, i.e., decoding pathway could start from \texttt{conv1} to \texttt{fc8}. Each macro-layer in AlexNet included exactly one convolutional or inner-product layer. 
We trained the decoding pathway with the encoding/classification pathway fixed. 

As shown in Figure~\ref{fig:alex-recon1}, the images reconstructed from any layer, even including the top 1000-way classification layer, were almost visually perfect.\footnote{For the \texttt{fc6} and \texttt{fc7} layers, we applied inner-product followed by relu nonlinearity; for the \texttt{fc8} layer, we applied only inner-product, but not softmax nonlinearity.} 
Only the local contrast and color saturation became slightly different from the original images as the layer went higher. 
The surprisingly good reconstruction quality suggests that the features produced by AlexNet preserved nearly all the information of the input except for the spatial invariance gained by the max-pooling layers. 

As commonly believed, learning task-relevant features for classification and preserving information were conflicting to some extent, since the ``nuisance'' should be removed for supervised tasks. 
According to our experiments, the locational details in different scales were almost the only information significantly neutralized by the deep neural network. 
For the convolutional and inner-product layers, it seems important to encode the input into a better (e.g., task-relevant) form without information loss. 

We conducted similar experiments based on the 16-layer VGGNet. 
As no results using the unpooling with fixed switches had been reported yet, we trained the decoding pathways for both SAE-first (with fixed unpooling switches) and SWWAE-first (with known unpooling switches).
We described the detailed training strategy in Section~\ref{sec:cls}. 
In Figure~\ref{fig:vgg-recon1}, we showed the reconstruction examples up to the 5\textsuperscript{th} macro-layer (the 13\textsuperscript{th} layer). 
Images reconstructed by SAE-first were blurry for higher layers. 
In contrast, SWWAE-first could well recover the shape details from the \texttt{pool5} features. 
In addition, the SWWAE-first model could also reasonably reconstruct non-ImageNet and even non-natural images like text screenshots, depth maps, and cartoon pictures, as shown in \suppx{sub:more-image-recon}{Figure~\ref{fig:vgg-recon-extra}}. These results suggest that the high-level feature representations were also adaptable to other domains. 

Since the architecture was much deeper than AlexNet, VGGNet resulted in noisier reconstruction. 
Assuming the ability of preserving information as a helpful property for deep neural network, we took the reconstruction loss as an auxiliary objective function for training the classification network, as will be described in Section~\ref{sec:cls}. 

\cutsectionbefore
\subsection{Image classification with augmented architectures \label{sec:cls}}
\cutsectionafter

We took as the baseline the 16-layer VGGNet (\citet{vggnet}'s Model D), one of the best open source convolutional neural networks for large-scale image classification.  

We needed only to use the classification pathway for testing. 
We report results with the following two schemes for sampling patches to show both more ablative and more practical performance on single networks. 
\begin{description}\cutitem
\item[Single-crop] We resized the test image, making its shorter edge $256$ pixels, and used only the single $224\times224$ patch (without mirroring) at the center to compute the classification score. 
It allowed us to examine the tradeoff between training and validation performance without complicated post-processing. \cutitem
\item[Convolution] We took the VGGNet as a fully convolutional network and used a global average-pooling to fuse the classification scores obtained at different locations in the grid. 
The test image was resized to $256$ pixels for the shorter edge and mirrored to go through the convolution twice. 
It was a replication of Section~3.2 of \citep{vggnet}.
\cutitem
\end{description}

We report the experimental results in Table~\ref{tab:reg-cls-error}.
Several VGGNet (classification pathway only) results are presented to justify the validity of our baseline implementation.  
As a replication of  \citet{vggnet}'s ``single-scale'' method, our second post-processing scheme could achieve similar comparable accuracy. 
Moreover, finetuning the pretrained VGGNet model further without the augmented decoding network using the same training procedure did not lead to significant performance change. 

\begin{table*}[!t]

{ \small{} 

\begin{centering}
\begin{tabular}{c|>{\centering}p{0.07\textwidth}>{\centering}p{0.07\textwidth}|>{\centering}p{0.07\textwidth}>{\centering}p{0.07\textwidth}|>{\centering}p{0.07\textwidth}|>{\centering}p{0.07\textwidth}}
\hline 
Sampling & \multicolumn{4}{c|}{Single-crop (center patch, no mirroring)} & \multicolumn{2}{c}{Convolution}\tabularnewline
\hline 
Errors & \multicolumn{2}{c|}{Top-1} & \multicolumn{2}{c|}{Top-5} & \multicolumn{1}{c|}{Top-1} & \multicolumn{1}{c}{Top-5}\tabularnewline
\hline 
Model & Train & Val. & Train & Val. & \multicolumn{2}{c}{Validation}\tabularnewline
\hline 
\hline 
VGGNet $^{\dagger}$ & -- & -- & -- & -- & 27.0\hspace{1bp}$^{*}$  & 8.8\hspace{1bp}$^{*}$ \tabularnewline
VGGNet $^{\dagger}$ & -- & -- & -- & -- & 26.8\hspace{1bp}$^{**}\hspace{-4bp}$  & 8.7\hspace{1bp}$^{**}\hspace{-4bp}$ \tabularnewline
\hline 
VGGNet & 17.43 & 29.05 & 4.02 & 10.07 & 26.97 & 8.94\tabularnewline
\hline 
SAE-first & 15.36 & 27.70 & 3.13 & \textcolor{white}{0}9.28 & 26.09 & 8.30\tabularnewline
SAE-all & 15.64 & 27.54 & 3.23 & \textcolor{white}{0}9.17 & 26.10 & 8.21\tabularnewline
SAE-layerwise & 16.20 & 27.60 & 3.42 & \textcolor{white}{0}9.19 & 26.06 & 8.17\tabularnewline
SWWAE-first & \textbf{15.10} & 27.60 & \textbf{3.08} & \textcolor{white}{0}9.23 & 25.87 & 8.14\tabularnewline
SWWAE-all & 15.67 & \textbf{27.39} & 3.24 & \textbf{\textcolor{white}{0}}\textbf{9.06} & \textbf{25.79} & \textbf{8.13}\tabularnewline
SWWAE-layerwise & 15.42 & 27.53 & 3.32 & \textcolor{white}{0}9.10 & 25.97 & 8.20\tabularnewline
\hline 
\end{tabular}
\par\end{centering}

\begin{centering}

\par\end{centering}

{ \footnotesize{} 

\begin{centering}
\smallskip{}
$^{\dagger}$ The numbers in the last rows are from Table~3 (Model D) in \citet{vggnet} (the most comparable to our settings).\protect\footnotemark
\par\end{centering}

\begin{centering}
$^{*}$~from a slightly different model trained with
single-scale ($256$px) data augmentation. $^{**}$~Test
scale is $384$px. 
\par\end{centering}

 }  

 } 
 
\centering{}
\cutcaptionabove
\vspace*{-0.1in}
\caption{\label{tab:reg-cls-error} Classification errors on ImageNet ILSVRC-2012
validation dataset based on 16-layer VGGNet. SAE models use the unpooling
with fixed switches, and SWWAE models uses the unpooling with known
switches. }
\cutfigurebelow 
\end{table*}

As a general trend, all of the networks augmented with autoencoders outperformed the baseline VGGNet by a noticeable margin. 
In particular, compared to the VGGNet baseline, the SWWAE-all model reduced the top-1 errors by $1.66\%$  and $1.18\%$ for the single-crop and convolution schemes, respectively. 
It also reduced the top-5 errors by $1.01\%$ and $0.81\%$, which are $10\%$ and $9\%$ relative to the baseline errors. 

To the best of our knowledge, this work provides the first experimental results to demonstrate the effectiveness of unsupervised learning objectives for improving the state-of-the-art image classification performance on large-scale realistic datasets. 
For SWWAE-all, the validation accuracy in Table 1 was achieved in $\sim$16 epochs, which took 4\textasciitilde{}5 days on a workstation with 4 Nvidia Titan X GPUs. 
Taking pretrained VGGNet as the reference, 75\% of the relative accuracy improvement ($\sim$1.25\% absolute top-1 accuracy improvement) could be achieved in $\sim$4 epochs ($\sim$1 day).

Apart from the general performance gain due to reconstructive decoding pathways, the architecture changes could result in relatively small differences.  
Compared to SWWAE-layerwise, SWWAE-all led to slightly higher accuracy, suggesting the usefulness of posing a higher requirement on the top convolutional features for preserving the input information.
The slight performance gain of SWWAE-all over SAE-all with fixed unpooling switches indicates that the switch connections could alleviate the difficulty of learning a stacked convolutional autoencoder. 
In the meanwhile, it also suggests that, without pooling switches, the decoding pathway can benefit the classification network learning similarly.
Using the unpooling with fixed switches, the decoding pathway may not be limited for reconstruction, but can also be designed for the structured outputs that are not locationally aligned with the input images (e.g, adjacent frames in videos, another viewpoint of the input object). 

\footnotetext{In our experiments, the 16-layer VGGNet (\citet{vggnet}'s Model D) achieved 10.07\% for the single-crop scheme and 8.94\% for the convolution scheme (in a single scale), which is comparable to 8.8\% in Table~3 of \citep{vggnet}. In that table, the best reported number for the Model D was 8.1\%, but it is trained and tested using a different resizing and cropping method, thus not comparable to our results.}

To figure out whether the performance gain was due to the potential regularization effects of the decoding pathway or not, we evaluated the networks on 50,000 images randomly chosen from the training set. 
Interestingly, the networks augmented with autoencoders achieved lower training errors than the baseline VGGNet. 
Hence, rather than regularizing, it is more likely that the auxiliary unsupervised loss helped the CNN to find better local optima in supervised learning.  
Compared to SAE/SWWAE-all, SAE/SWWAE-first led to lower training errors but higher validation errors, a typical symptom of slight overfitting. 
Thus, incorporating layer-wise reconstruction loss was an effective way to regularize the network training.

We provide more discussion for the decoding pathways in \supp{sec:more-exp}, including image reconstruction results after finetuning the augmented networks\suppd{sub:more-image-recon}, training curves\suppd{sub:training-curve}, and comparison between the pretrained and finetuned convolution filters\suppd{sub:filters}.
 
\cutsectionbefore
\section{Conclusion}
\cutsectionafter

We proposed a simple and effective way to incorporate unsupervised objectives into large-scale classification network learning by augmenting the existing network with reconstructive decoding pathways. 
Using the resultant autoencoder for image reconstruction, we demonstrated the ability of preserving input information by intermediate representation as an important property of modern deep neural networks trained for large-scale image classification. 
We leveraged this property further by training the augmented network composed of both the classification and decoding pathways.
This method improved the performance of the 16-layer VGGNet, one of the best existing networks for image classification by a noticeable margin. 
We investigated different variants of the autoencoder, and showed that 
1) the pooling switch connections between the encoding and decoding pathways were helpful, but not critical for improving the performance of the classification network in large-scale settings; 
2) the decoding pathways mainly helped the supervised objective reach a better optimum; and 
3) the layer-wise reconstruction loss could effectively regularize the solution to the joint objective. 
We hope this paper will inspire further investigations on the use of unsupervised learning in a large-scale setting.
 
\section*{Acknowledgements}

This work was funded by Software R{\&}D Center, Samsung Electronics Co., Ltd; ONR N00014-13-1-0762; and NSF CAREER IIS-1453651. 
We also thank NVIDIA for donating K40c and TITAN X GPUs.
We thank Jimei Yang, Seunghoon Hong, Ruben Villegas, Wenling Shang, Kihyuk Sohn, and other collaborators for helpful discussions. 
 
\bibliographystyle{icml2016}
\bibliography{ref}

\onecolumn

\appendix

\renewcommand{\thefigure}{A-\arabic{figure}}
\renewcommand{\thetable}{A-\arabic{table}}
\renewcommand{\theequation}{A-\arabic{equation}}
\renewcommand{\thealgorithm}{A-\arabic{algorithm}}
\renewcommand{\thesection}{A\arabic{section}}

\mbox{\Large{} \textbf{Appendices}\par}

\vspace{1em}

\setcounter{figure}{0}
\setcounter{table}{0}
\setcounter{equation}{0}
\setcounter{algorithm}{0}
\setcounter{section}{0}

\section{Parameters for VGGNet-based models}
\label{sec:vgg-param}

\begin{table}[H]
{ \small{} 
\renewcommand*\arraystretch{1.15}

\begin{centering}
\begin{tabular}{c|>{\centering}p{0.3\columnwidth}|>{\centering}p{0.3\columnwidth}}
\hline 
Macro- & Learning rate & Loss weighting \textsuperscript{1}\tabularnewline
\cline{2-3} 
layer & SAE-layerwise & SAE-layerwise/all\tabularnewline
\hline 
1 & $3\times10^{-9}$ & $1\times10^{-4}$\tabularnewline
\hline 
2 & $1\times10^{-8}$ & $1\times10^{-12}$\tabularnewline
\hline 
3 & $3\times10^{-12}$ & $1\times10^{-12}$\tabularnewline
\hline 
4 & $1\times10^{-12}$ & $1\times10^{-12}$\tabularnewline
\hline 
5 & $1\times10^{-11}$ & $1\times10^{-10}$\tabularnewline
\hline 
\end{tabular}
\par\end{centering}

\begin{centering}
{ \footnotesize{}  LR: learning rate; \textsuperscript{1} the top-level
softmax is weighted by $1$. }
\par\end{centering}

\renewcommand*\arraystretch{1}
} 

\caption{Layer-wise training parameters for networks augmented from VGGNet \label{tab:vgg-param}}
\end{table}

We report the learning parameters for 16-layer VGGNet-based model in Table~\ref{tab:vgg-param}. 
We chose the learning rates that lead to the largest decrease in the reconstruction loss in the first 2000 iterations for each layer. 
The ``loss weighting'' are balancing factors for reconstruction losses in different layers varied to make them comparable in magnitude.
In particular, we computed image reconstruction loss against RGB values normalized to [0,1], which are different in scale from intermediate features. 
We also did not normalize the reconstruction loss with feature dimensions for any layer.  

\FloatBarrier

\section{More experimental results and discussions}
\label{sec:more-exp}

\subsection{Learned filters \label{sub:filters}}

\begin{figure}[h]
\hfill{}\hfill{}
\subfloat[Pretrained VGGNet]{
\includegraphics[width=0.3\textwidth]{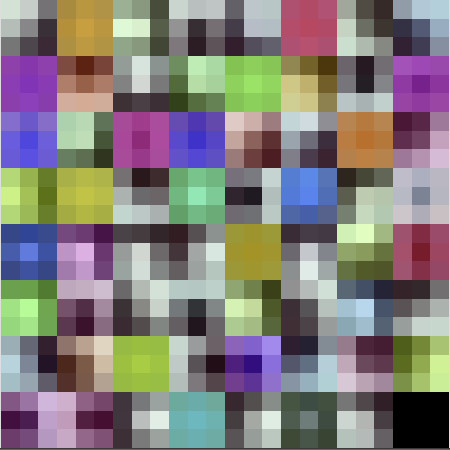}
}
\hfill{}
\subfloat[Finetuned SWWAE-all]{
\includegraphics[width=0.3\textwidth]{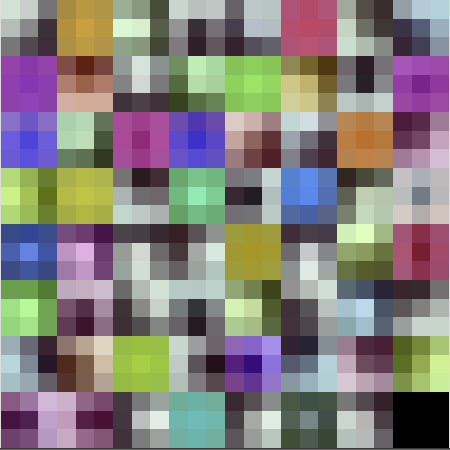}
}
\hfill{}\hfill{}
\par
\caption{Visualization of the normalizaed first-layer convolution filters in 16-layer VGGNet-based network. \label{fig:layer1-vis} The filters of the SWWAE-all model had nearly the same patterns to those of the pretrained VGGNet, but showed stronger contrast. It is more clear see the difference if displaying the two images alternatively in the same place. (online example: \url{http://www.ytzhang.net/files/publications/2016-icml-recon-dec/filters/})}

\end{figure}

Compared to the baseline VGGNet, the finetuned SWWAE-all model demonstrated $\sim 35\%$ element-wise relative change of the filter weights on average for all the layers.
A small portion of the filters showed stronger contrast after finetuning. Qualitatively, the finetuned filters kept the pretrained visual shapes. 
In Figure~\ref{fig:layer1-vis}, we visualize the first-layer $3\times 3$ convolution filters. 

\FloatBarrier

\subsection{Training curve \label{sub:training-curve}}

\begin{figure*}[!t]
\hfill{}
\includegraphics[width=.42\textwidth]{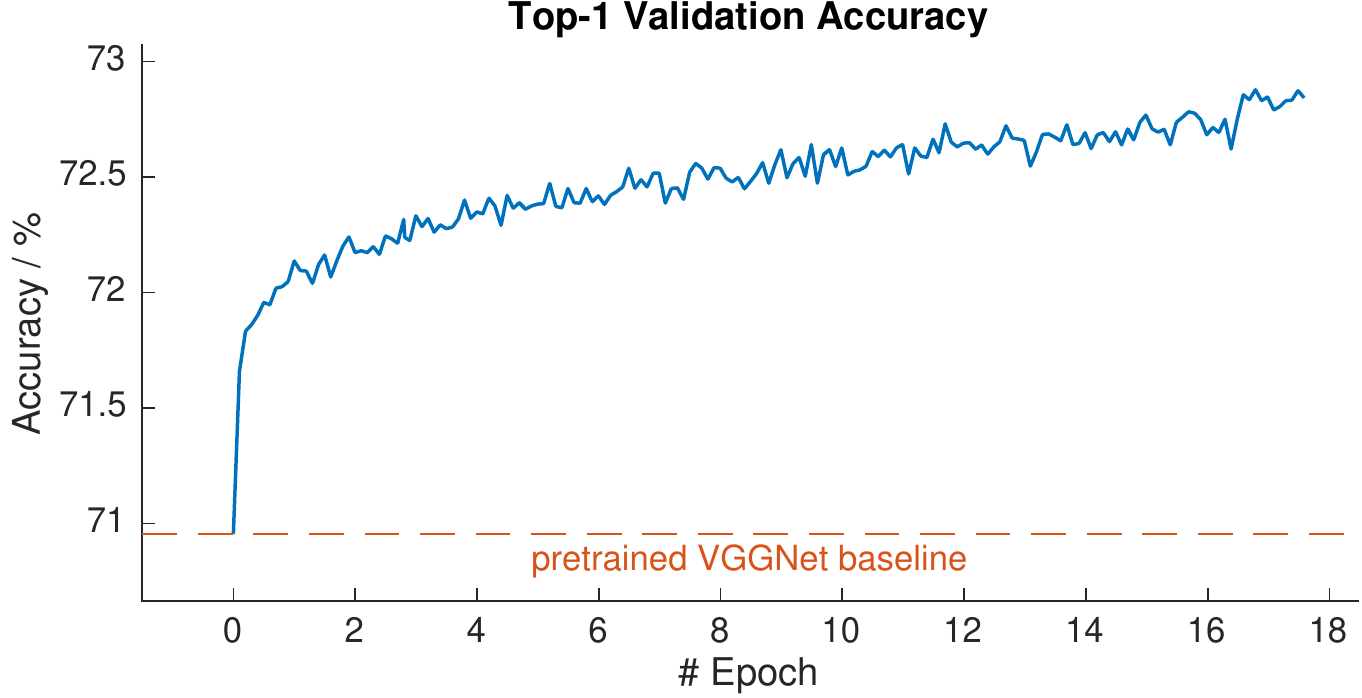}
\hfill{}
\includegraphics[width=.42\textwidth]{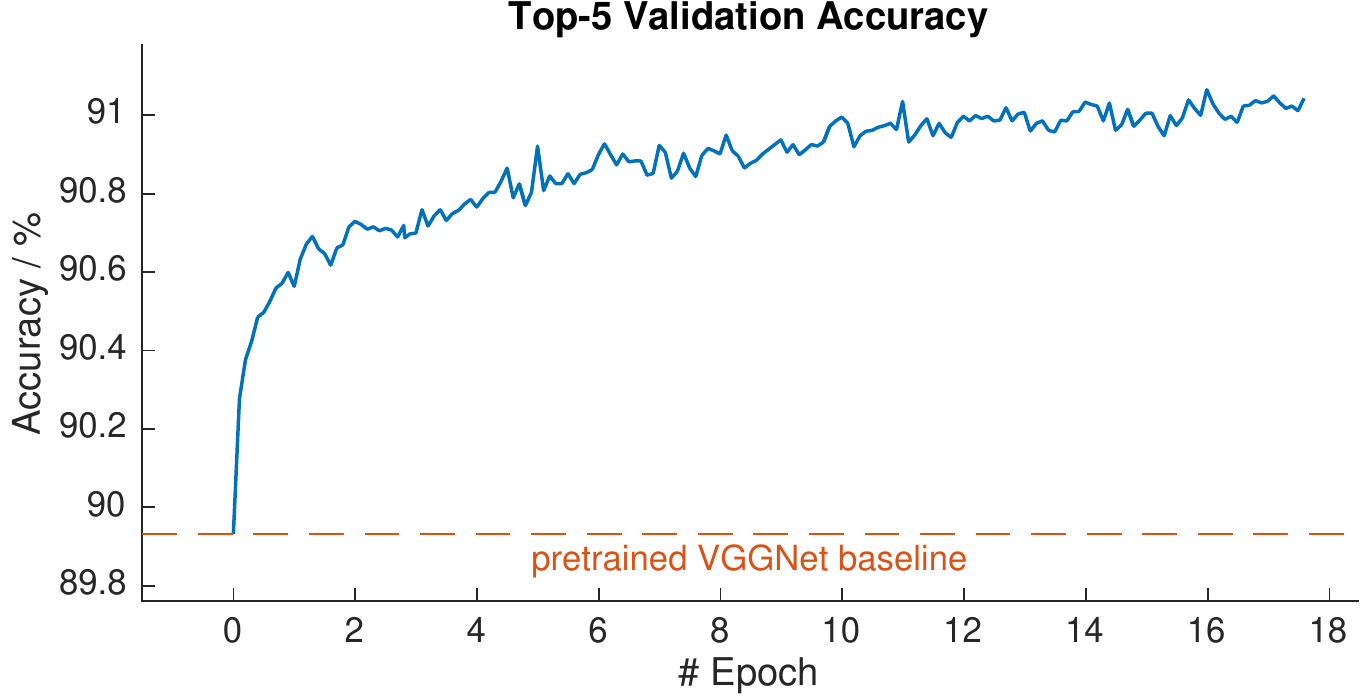}
\hfill{}

\caption{Training curves for the single-crop validation accuracy of VGGNet-based SWWAE-all models. \label{fig:traning-curve}
}
\end{figure*}

In Figure~\ref{fig:traning-curve}, we report the training curves of validation accuracy for SWWAE-all, where the pretrained VGGNet classification network and decoder network were taken as the starting point.

\subsection{Selection of different model variants \label{sub:model-selection}}
The performance for different variants of the augmented network are comparable, but we can still choose the best available one. In particular, we provide following discussions. 
\begin{itemize}
\item Since the computational costs were similar for training and the same for testing, we can use the best available architecture depending on tasks. 
For example, when using decoding pathways for spatially corresponded tasks like reconstruction (as in our paper) and segmentation, we can use the SWWAE. 
For more general objectives like predicting next frames, where pooling switches are non-transferrable, we can still use ordinary SAEs to get competitive performance. 
\item S(WW)AE-first has less hyper-parameters than S(WW)AE-all, and can be trained first for quick parameter search. It can be switched to *-all for better performance. 
\end{itemize}

\subsection{Ladder networks \label{sub:ladder}}

We tried training a ladder network following the same procedures of pretraining auxiliary pathways and finetuning the whole network as for our models, which is also similar to \citet{ladder-semi}'s strategy.
We used the augmented multi-layer perceptron (AMLP) combinator, which \citet{ladder-deconstruct} proposed as the best combinator function.
Different from the previous work conducted on the variants of MNIST dataset, the pretrained VGGNet does not have batch normalization (BN) layers, which pushed us to remove the BN layers from the ladder network.
However, BN turned out to be critical for proper noise injection, and the non-BN ladder network did not perform well. 
It might suggest that our models are easier to pair with a standard convolutional network and train on large-scale datasets. 

\FloatBarrier

\subsection{Image reconstruction\label{sub:more-image-recon}}

\begin{figure}

{ \small{}

\begin{centering}
\begin{tabular}{c|c|cc|cccccc}
\hline 
Model & L2 Loss & \multicolumn{2}{c|}{ImageNet} & \multicolumn{6}{c}{Non-ImageNet \textsuperscript{1}}\tabularnewline
\hline 
\begin{minipage}[t]{7em}\begin{center}
\vspace{1.2em}
Ground truth
\par\end{center}\end{minipage} & \begin{minipage}[t]{3em}\begin{center}
\vspace{1.5em}
-
\par\end{center}\end{minipage} & \hspace{-5pt}\begin{minipage}[t]{0.08\textwidth}\noindent \begin{center}
\vspace{-1em}
\includegraphics[width=1\columnwidth]{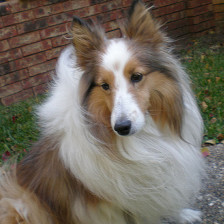}\vspace{2pt}
\par\end{center}
\end{minipage}\hspace{-5pt} & \hspace{-5pt}\begin{minipage}[t]{0.08\textwidth}
\noindent \begin{center}
\vspace{-1em}
\includegraphics[width=1\columnwidth]{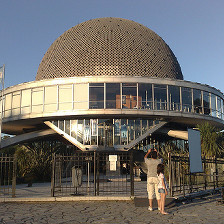}\vspace{2pt}

\par\end{center}
\end{minipage}\hspace{-5pt} & \hspace{-5pt}\begin{minipage}[t]{0.08\textwidth}
\noindent \begin{center}
\vspace{-1em}
\includegraphics[width=1\columnwidth]{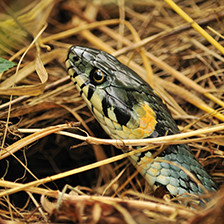}\vspace{2pt}

\par\end{center}
\end{minipage}\hspace{-5pt} & \hspace{-5pt}\begin{minipage}[t]{0.08\textwidth}
\noindent \begin{center}
\vspace{-1em}
\includegraphics[width=1\columnwidth]{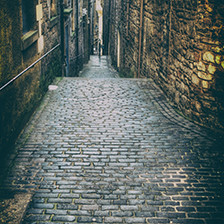}\vspace{2pt}

\par\end{center}
\end{minipage}\hspace{-5pt} & \hspace{-5pt}\begin{minipage}[t]{0.08\textwidth}
\noindent \begin{center}
\vspace{-1em}
\includegraphics[width=1\columnwidth]{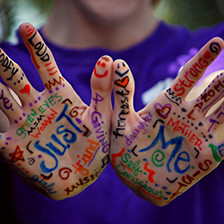}\vspace{2pt}

\par\end{center}
\end{minipage}\hspace{-5pt} & \hspace{-5pt}\begin{minipage}[t]{0.08\textwidth}
\noindent \begin{center}
\vspace{-1em}
\includegraphics[width=1\columnwidth]{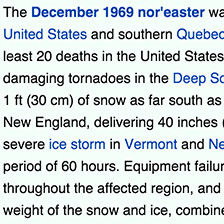}\vspace{2pt}

\par\end{center}
\end{minipage}\hspace{-5pt} & \hspace{-5pt}\begin{minipage}[t]{0.08\textwidth}
\noindent \begin{center}
\vspace{-1em}
\includegraphics[width=1\columnwidth]{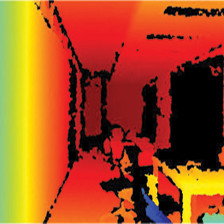}\vspace{2pt}

\par\end{center}
\end{minipage}\hspace{-5pt} & \hspace{-5pt}\begin{minipage}[t]{0.08\textwidth}
\noindent \begin{center}
\vspace{-1em}
\includegraphics[width=1\columnwidth]{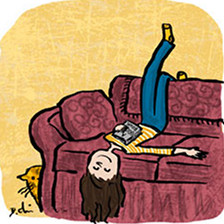}\vspace{2pt}

\par\end{center}
\end{minipage}\hspace{-5pt}\tabularnewline
\hline 
\begin{minipage}[t]{7em}\begin{center}
\vspace{0.5em}
SWWAE-first (Pretrained, fixing encoder)
\par\end{center}\end{minipage} & \begin{minipage}[t]{3em}\begin{center}
\vspace{1.5em}
513.4
\par\end{center}\end{minipage} & \hspace{-5pt}\begin{minipage}[t]{0.08\textwidth}
\noindent \begin{center}
\vspace{-1em}
\includegraphics[width=1\columnwidth]{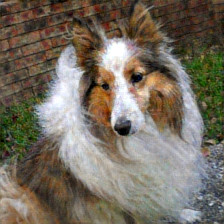}
\par\end{center}
\end{minipage}\hspace{-5pt} & \hspace{-5pt}\begin{minipage}[t]{0.08\textwidth}
\noindent \begin{center}
\vspace{-1em}
\includegraphics[width=1\columnwidth]{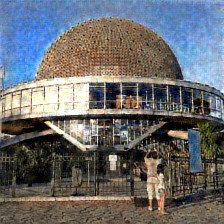}
\par\end{center}
\end{minipage}\hspace{-5pt} & \hspace{-5pt}\begin{minipage}[t]{0.08\textwidth}
\noindent \begin{center}
\vspace{-1em}
\includegraphics[width=1\columnwidth]{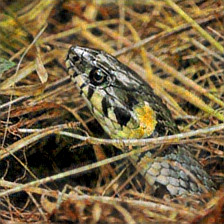}
\par\end{center}
\end{minipage}\hspace{-5pt} & \hspace{-5pt}\begin{minipage}[t]{0.08\textwidth}
\noindent \begin{center}
\vspace{-1em}
\includegraphics[width=1\columnwidth]{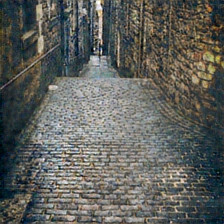}
\par\end{center}
\end{minipage}\hspace{-5pt} & \hspace{-5pt}\begin{minipage}[t]{0.08\textwidth}
\noindent \begin{center}
\vspace{-1em}
\includegraphics[width=1\columnwidth]{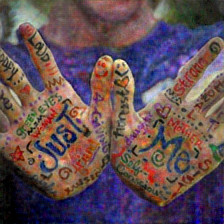}
\par\end{center}
\end{minipage}\hspace{-5pt} & \hspace{-5pt}\begin{minipage}[t]{0.08\textwidth}
\noindent \begin{center}
\vspace{-1em}
\includegraphics[width=1\columnwidth]{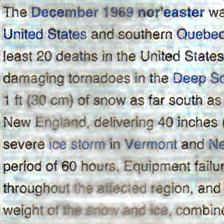}
\par\end{center}
\end{minipage}\hspace{-5pt} & \hspace{-5pt}\begin{minipage}[t]{0.08\textwidth}
\noindent \begin{center}
\vspace{-1em}
\includegraphics[width=1\columnwidth]{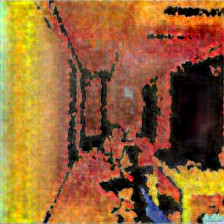}
\par\end{center}
\end{minipage}\hspace{-5pt} & \hspace{-5pt}\begin{minipage}[t]{0.08\textwidth}
\noindent \begin{center}
\vspace{-1em}
\includegraphics[width=1\columnwidth]{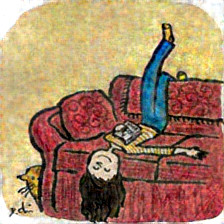}
\par\end{center}
\end{minipage}\hspace{-5pt}\tabularnewline
\begin{minipage}[t]{7em}\begin{center}
\vspace{0.5em}
SWWAE-first (Finetuned with encoder)
\par\end{center}\end{minipage} & \begin{minipage}[t]{3em}\begin{center}
\vspace{1.5em}
462.2
\par\end{center}\end{minipage} & \hspace{-5pt}\begin{minipage}[t]{0.08\textwidth}
\noindent \begin{center}
\vspace{-1em}
\includegraphics[width=1\columnwidth]{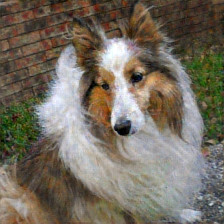}\vspace{2pt}

\par\end{center}
\end{minipage}\hspace{-5pt} & \hspace{-5pt}\begin{minipage}[t]{0.08\textwidth}
\noindent \begin{center}
\vspace{-1em}
\includegraphics[width=1\columnwidth]{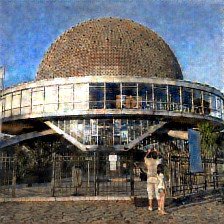}\vspace{2pt}

\par\end{center}
\end{minipage}\hspace{-5pt} & \hspace{-5pt}\begin{minipage}[t]{0.08\textwidth}
\noindent \begin{center}
\vspace{-1em}
\includegraphics[width=1\columnwidth]{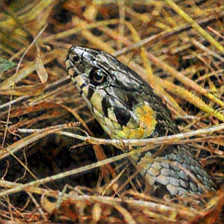}\vspace{2pt}

\par\end{center}
\end{minipage}\hspace{-5pt} & \hspace{-5pt}\begin{minipage}[t]{0.08\textwidth}
\noindent \begin{center}
\vspace{-1em}
\includegraphics[width=1\columnwidth]{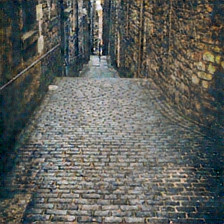}\vspace{2pt}

\par\end{center}
\end{minipage}\hspace{-5pt} & \hspace{-5pt}\begin{minipage}[t]{0.08\textwidth}
\noindent \begin{center}
\vspace{-1em}
\includegraphics[width=1\columnwidth]{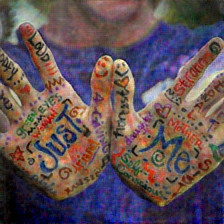}\vspace{2pt}

\par\end{center}
\end{minipage}\hspace{-5pt} & \hspace{-5pt}\begin{minipage}[t]{0.08\textwidth}
\noindent \begin{center}
\vspace{-1em}
\includegraphics[width=1\columnwidth]{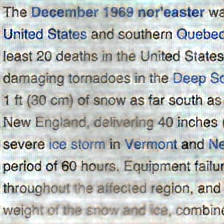}\vspace{2pt}

\par\end{center}
\end{minipage}\hspace{-5pt} & \hspace{-5pt}\begin{minipage}[t]{0.08\textwidth}
\noindent \begin{center}
\vspace{-1em}
\includegraphics[width=1\columnwidth]{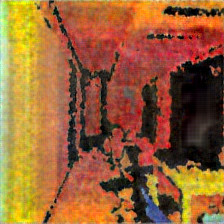}\vspace{2pt}

\par\end{center}
\end{minipage}\hspace{-5pt} & \hspace{-5pt}\begin{minipage}[t]{0.08\textwidth}
\noindent \begin{center}
\vspace{-1em}
\includegraphics[width=1\columnwidth]{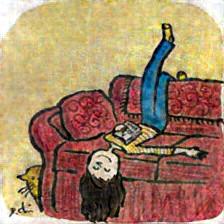}\vspace{2pt}

\par\end{center}
\end{minipage}\hspace{-5pt}\tabularnewline
\begin{minipage}[t]{7em}\begin{center}
\vspace{0.5em}
SWWAE-all (Finetuned with encoder)
\par\end{center}\end{minipage} & \begin{minipage}[t]{3em}\begin{center}
\vspace{1.5em}
493.0
\par\end{center}\end{minipage} & \hspace{-5pt}\begin{minipage}[t]{0.08\textwidth}\noindent \begin{center}
\vspace{-1em}
\includegraphics[width=1\columnwidth]{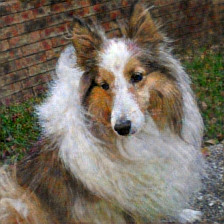}\vspace{2pt}
\par\end{center}
\end{minipage}\hspace{-5pt} & \hspace{-5pt}\begin{minipage}[t]{0.08\textwidth}\noindent \begin{center}
\vspace{-1em}
\includegraphics[width=1\columnwidth]{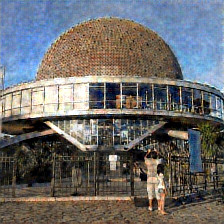}\vspace{2pt}
\par\end{center}
\end{minipage}\hspace{-5pt} & \hspace{-5pt}\begin{minipage}[t]{0.08\textwidth}\noindent \begin{center}
\vspace{-1em}
\includegraphics[width=1\columnwidth]{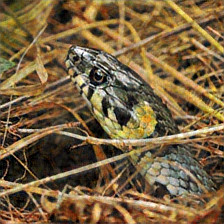}\vspace{2pt}
\par\end{center}
\end{minipage}\hspace{-5pt} & \hspace{-5pt}\begin{minipage}[t]{0.08\textwidth}\noindent \begin{center}
\vspace{-1em}
\includegraphics[width=1\columnwidth]{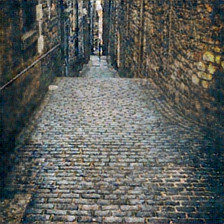}\vspace{2pt}
\par\end{center}
\end{minipage}\hspace{-5pt} & \hspace{-5pt}\begin{minipage}[t]{0.08\textwidth}\noindent \begin{center}
\vspace{-1em}
\includegraphics[width=1\columnwidth]{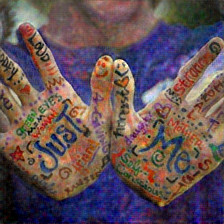}\vspace{2pt}
\par\end{center}
\end{minipage}\hspace{-5pt} & \hspace{-5pt}\begin{minipage}[t]{0.08\textwidth}\noindent \begin{center}
\vspace{-1em}
\includegraphics[width=1\columnwidth]{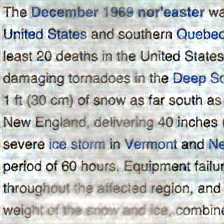}\vspace{2pt}
\par\end{center}
\end{minipage}\hspace{-5pt} & \hspace{-5pt}\begin{minipage}[t]{0.08\textwidth}\noindent \begin{center}
\vspace{-1em}
\includegraphics[width=1\columnwidth]{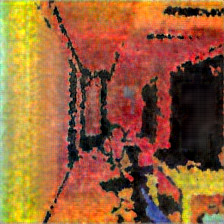}\vspace{2pt}
\par\end{center}
\end{minipage}\hspace{-5pt} & \hspace{-5pt}\begin{minipage}[t]{0.08\textwidth}\noindent \begin{center}
\vspace{-1em}
\includegraphics[width=1\columnwidth]{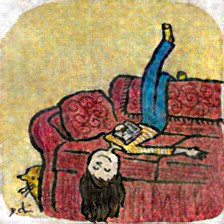}\vspace{2pt}
\par\end{center}
\end{minipage}\hspace{-5pt}\tabularnewline
\hline 
\end{tabular}
\par\end{centering}
\begin{centering}
\par\end{centering}

\begin{centering}
\begin{minipage}[t]{0.95\textwidth}\begin{flushleft}
\textsuperscript{1} The first three images are from morguefile.com;
the fourth is a screenshot of Wikipedia; the fifth is a depth image
from NYU dataset; the last is used with permission from Debbie Ridpath
Ohi at Inkygirl.com
\par\end{flushleft}\end{minipage}
\par\end{centering}

} 
 
\caption{\label{fig:vgg-recon-extra}Image reconstruction from pool5 features to
images. The reconstruction loss is computed on the ILSVRC2012 validation
set and measured with L2-distance with the ground truth (RGB values
are in $[0,1]$). The first 2 example images are from the ILSVRC2012
validation set (excluding the 100 categories). The rest are not in
ImageNet. }
\end{figure}

In Figure~\ref{fig:vgg-recon-extra}, we visualize the images reconstructed by the pretrained decoder of SWWAE-first and the final models for SWWAE-first/all, and reported the L2 reconstruction loss on the validation set. 
Finetuning the entire networks also resulted in better reconstruction quality, which is consistent with our assumption that enhancing the ability
of preserving input information can lead to better features for image classification. 
Since the shape details had already been well recovered by the pretrained decoder, the finetuned SWWAE-first/all mainly improved the accuracy of colors. Note that the decoder learning is more difficult for SWWAE-all than SWWAE-first, which explains its slightly higher reconstruction loss and better regularization ability. 

In Figure~\ref{fig:alex-recon2} and \ref{fig:vgg-recon2}, we showed more examples for reconstructing input images from pretrained neural network features for AlexNet and VGGNet. 

\clearpage

\newcommand{\alexrecontag}{\hspace{-5pt}\begin{minipage}[t]{6.5em}\begin{center}
\vspace{-0.3em}
SWWAE-first 
(\textbf{known} unpooling switches)
\par\end{center}\end{minipage}}

\begin{figure}[p]

{ \small{} 

\begin{centering}
\begin{tabular}{c|ccccccccc}
\hline 
Layer & image & pool1 & pool2 & conv3 & conv4 & pool5 & fc6 & fc7 & fc8\tabularnewline
\hline 

 \end{tabular}
\par\end{centering}

 }

\caption{\label{fig:alex-recon2}AlexNet reconstruction on ImageNet
ILSVRC2012 validation set.  (Best viewed when zoomed in on a screen.)}
\end{figure}

\clearpage
 
\providecommand{\vggfigcut}{\vspace{-1.5em}}
\providecommand{\vggfigcnp}{\vggfigcut\begin{centering}{\small{} (continued on next page) } \par \end{centering}}
\vfill

{ \small{} 

\begin{centering}
\begin{tabular}{c|cccccc}
\hline 
Layer & image & pool1 & pool2 & pool3 & pool4 & pool5\tabularnewline
\hline 
 \begin{minipage}[t]{8em}\begin{center}
\vspace{0em}
SAE-first
(\textbf{fixed} unpooling switches)
\par\end{center}\end{minipage} & \hspace{-5pt}\begin{minipage}[t]{0.12\textwidth}\noindent \begin{center}
\vspace{-1em}
\includegraphics[width=1\columnwidth]{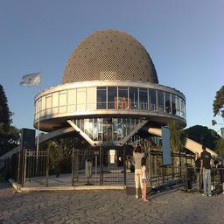}\vspace{2pt}
\par\end{center}
\end{minipage}\hspace{-5pt} & \hspace{-5pt}\begin{minipage}[t]{0.12\textwidth}\noindent \begin{center}
\vspace{-1em}
\includegraphics[width=1\columnwidth]{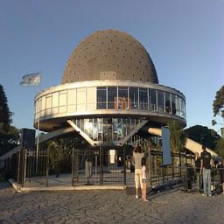}\vspace{2pt}
\par\end{center}
\end{minipage}\hspace{-5pt} & \hspace{-5pt}\begin{minipage}[t]{0.12\textwidth}\noindent \begin{center}
\vspace{-1em}
\includegraphics[width=1\columnwidth]{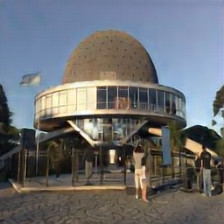}\vspace{2pt}
\par\end{center}
\end{minipage}\hspace{-5pt} & \hspace{-5pt}\begin{minipage}[t]{0.12\textwidth}\noindent \begin{center}
\vspace{-1em}
\includegraphics[width=1\columnwidth]{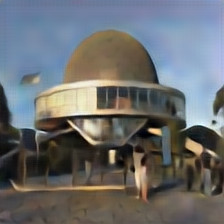}\vspace{2pt}
\par\end{center}
\end{minipage}\hspace{-5pt} & \hspace{-5pt}\begin{minipage}[t]{0.12\textwidth}\noindent \begin{center}
\vspace{-1em}
\includegraphics[width=1\columnwidth]{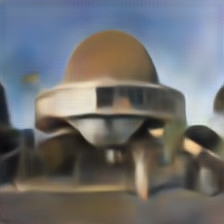}\vspace{2pt}
\par\end{center}
\end{minipage}\hspace{-5pt} & \hspace{-5pt}\begin{minipage}[t]{0.12\textwidth}\noindent \begin{center}
\vspace{-1em}
\includegraphics[width=1\columnwidth]{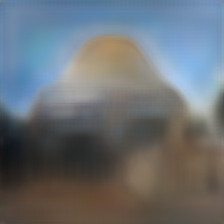}\vspace{2pt}
\par\end{center}
\end{minipage}\hspace{-5pt}\tabularnewline
\hline
 \end{tabular}
\par\end{centering}

 }
 \vspace{-1.5em}
\vggfigcnp
\vfill \clearpage

\vfill

{ \small{} 

\begin{centering}
\begin{tabular}{c|cccccc}
\hline 
Layer & image & pool1 & pool2 & pool3 & pool4 & pool5\tabularnewline
\hline 
 \begin{minipage}[t]{8em}\begin{center}
\vspace{0em}
SAE-first
(\textbf{fixed} unpooling switches)
\par\end{center}\end{minipage} & \hspace{-5pt}\begin{minipage}[t]{0.12\textwidth}\noindent \begin{center}
\vspace{-1em}
\includegraphics[width=1\columnwidth]{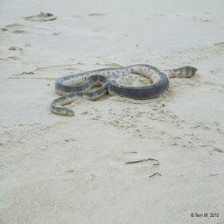}\vspace{2pt}
\par\end{center}
\end{minipage}\hspace{-5pt} & \hspace{-5pt}\begin{minipage}[t]{0.12\textwidth}\noindent \begin{center}
\vspace{-1em}
\includegraphics[width=1\columnwidth]{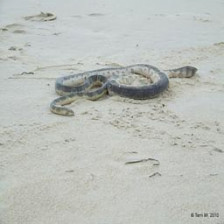}\vspace{2pt}
\par\end{center}
\end{minipage}\hspace{-5pt} & \hspace{-5pt}\begin{minipage}[t]{0.12\textwidth}\noindent \begin{center}
\vspace{-1em}
\includegraphics[width=1\columnwidth]{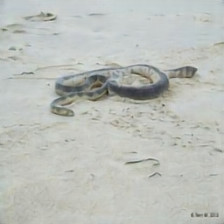}\vspace{2pt}
\par\end{center}
\end{minipage}\hspace{-5pt} & \hspace{-5pt}\begin{minipage}[t]{0.12\textwidth}\noindent \begin{center}
\vspace{-1em}
\includegraphics[width=1\columnwidth]{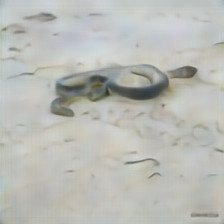}\vspace{2pt}
\par\end{center}
\end{minipage}\hspace{-5pt} & \hspace{-5pt}\begin{minipage}[t]{0.12\textwidth}\noindent \begin{center}
\vspace{-1em}
\includegraphics[width=1\columnwidth]{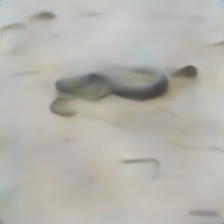}\vspace{2pt}
\par\end{center}
\end{minipage}\hspace{-5pt} & \hspace{-5pt}\begin{minipage}[t]{0.12\textwidth}\noindent \begin{center}
\vspace{-1em}
\includegraphics[width=1\columnwidth]{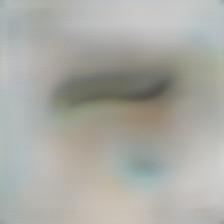}\vspace{2pt}
\par\end{center}
\end{minipage}\hspace{-5pt}\tabularnewline
\hline
 \end{tabular}
\par\end{centering}

 }
 \vggfigcnp
\vfill \clearpage

\vfill

{ \small{} 

\begin{centering}
\begin{tabular}{c|cccccc}
\hline 
Layer & image & pool1 & pool2 & pool3 & pool4 & pool5\tabularnewline
\hline 
 \begin{minipage}[t]{8em}\begin{center}
\vspace{0em}
SAE-first
(\textbf{fixed} unpooling switches)
\par\end{center}\end{minipage} & \hspace{-5pt}\begin{minipage}[t]{0.12\textwidth}\noindent \begin{center}
\vspace{-1em}
\includegraphics[width=1\columnwidth]{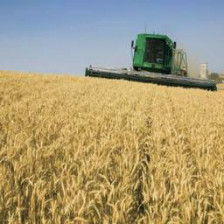}\vspace{2pt}
\par\end{center}
\end{minipage}\hspace{-5pt} & \hspace{-5pt}\begin{minipage}[t]{0.12\textwidth}\noindent \begin{center}
\vspace{-1em}
\includegraphics[width=1\columnwidth]{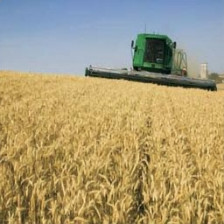}\vspace{2pt}
\par\end{center}
\end{minipage}\hspace{-5pt} & \hspace{-5pt}\begin{minipage}[t]{0.12\textwidth}\noindent \begin{center}
\vspace{-1em}
\includegraphics[width=1\columnwidth]{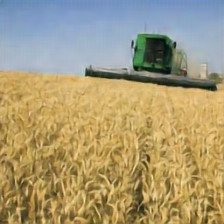}\vspace{2pt}
\par\end{center}
\end{minipage}\hspace{-5pt} & \hspace{-5pt}\begin{minipage}[t]{0.12\textwidth}\noindent \begin{center}
\vspace{-1em}
\includegraphics[width=1\columnwidth]{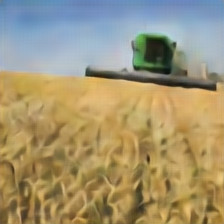}\vspace{2pt}
\par\end{center}
\end{minipage}\hspace{-5pt} & \hspace{-5pt}\begin{minipage}[t]{0.12\textwidth}\noindent \begin{center}
\vspace{-1em}
\includegraphics[width=1\columnwidth]{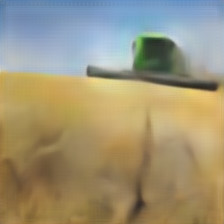}\vspace{2pt}
\par\end{center}
\end{minipage}\hspace{-5pt} & \hspace{-5pt}\begin{minipage}[t]{0.12\textwidth}\noindent \begin{center}
\vspace{-1em}
\includegraphics[width=1\columnwidth]{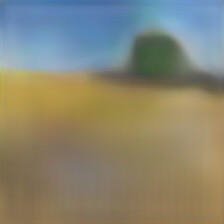}\vspace{2pt}
\par\end{center}
\end{minipage}\hspace{-5pt}\tabularnewline
\hline
 \end{tabular}
\par\end{centering}

 }
 \vspace{-2em}

\begin{figure}[H]
\vggfigcut
\vspace{-0.5em}
\hfill
\caption{\label{fig:vgg-recon2}VGGNet reconstruction on ImageNet ILSVRC2012 validation set. (Best viewed when zoomed in on a screen.)}
\end{figure}

\vfill \clearpage

\end{document}